%% file: main.tex
\documentclass[conference]{IEEEtran}
\IEEEoverridecommandlockouts

\usepackage[cmex10]{amsmath}
\usepackage{cite}
\usepackage[table]{xcolor}

\usepackage{amssymb, amsmath, amsfonts}
\usepackage{pifont}

\usepackage{enumitem}
\usepackage{path}
\usepackage{verbatim}
\usepackage{algorithm}
\usepackage{algorithmic}
\usepackage{framed}
\usepackage{footnote}
\usepackage{color}
\usepackage{cases}
\usepackage{subfigure}
\usepackage{theorem}
\usepackage{xurl}
\usepackage{acro}

\usepackage{tabularx}
\usepackage{booktabs}
\usepackage[colorlinks=true, allcolors=black]{hyperref}
\usepackage{threeparttable}

\usepackage{graphicx}

\usepackage{pgfplots}
\usepackage{pgfplotstable}
\pgfplotsset{compat=newest}
\graphicspath{{graphics/}}

{ \theorembodyfont{\normalfont} 

}

\newcounter{enumctr}

\DeclareFontFamily{U}{mathx}{\hyphenchar\font45}
\DeclareFontShape{U}{mathx}{m}{n}{<-> mathx10}{}
\DeclareSymbolFont{mathx}{U}{mathx}{m}{n}
\DeclareMathAccent{\widebar}{0}{mathx}{"73}

\DeclareAcronym{ann}{
    short = ANN,
    long  = Artificial Neural Networks,
}
\DeclareAcronym{bev}{
    short = BEV,
    long  = Battery Electric Vehicle,
}
\DeclareAcronym{cnn}{
    short = CNN,
    long  = Convolutional Neural Networks,
}
\DeclareAcronym{ev}{
    short = EV,
    long  = Electric Vehicle,
}
\DeclareAcronym{avg}{
    short = FedAvg,
    long  = Federated Averaging,
}
\DeclareAcronym{per}{
    short = FedPer,
    long  = Federated Personalization,
}
\DeclareAcronym{prox}{
    short = FedProx,
    long  = Federated Proximal,
}
\DeclareAcronym{rep}{
    short = FedRep,
    long  = Federated Representation Learning,
}
\DeclareAcronym{sgd}{
    short = FedSGD,
    long  = Federated Stochastic Gradient Descent,
}
\DeclareAcronym{fl}{
    short = FL,
    long  = Federated Learning,
}
\DeclareAcronym{gru}{
    short = GRU,
    long  = Gated Recurrent Unit,
}
\DeclareAcronym{itr}{
    short = ITR,
    long  = Iterations,
}
\DeclareAcronym{lstm}{
    short = LSTM,
    long  = Long Short-Term Memory,
}
\DeclareAcronym{mae}{
    short = MAE,
    long  = Mean Absolute Error,
}
\DeclareAcronym{ml}{
    short = ML,
    long  = Machine Learning,
}
\DeclareAcronym{rf}{
    short = RF,
    long  = Random Forest,
}
\DeclareAcronym{ts}{
    short = TS,
    long  = Time Step,
}
\DeclareAcronym{ved}{
    short = VED,
    long  = Vehicle Energy Dataset,
}
\DeclareAcronym{xgb}{
    short = XGB,
    long  = eXtreme Gradient Boosting,
}

\acsetup{
  list/template = description
}

\setlist[description]{align=left, leftmargin=2cm, style=nextline}

\newcolumntype{C}{>{\centering\arraybackslash}X} 
\begin{document}

    \title{\LARGE \bf Privacy-Aware Energy Consumption Modeling of Connected Battery Electric Vehicles using Federated Learning}
    
    \author{Sen Yan, \textit{Graduate Student Member, IEEE}, Hongyuan Fang, Ji Li, \textit{Member, IEEE}, \\Tomas Ward, \textit{Senior Member, IEEE}, Noel O'Connor, \textit{Member, IEEE}, and Mingming Liu, \textit{Senior Member, IEEE} \thanks{Manuscript received May 6, 2023; revised July 28, 2023 and October 20, 2023; accepted December 4, 2023. This work was supported by Science Foundation Ireland 21/FFP-P/10266 and 12/RC/2289\_P2 at Insight the SFI Research Centre for Data Analytics at Dublin City University. The dataset is now available on GitHub and can be accessed via {\tt \url{ https://github.com/SFIEssential/FedBEV}}. (Corresponding authors: Ji Li; Mingming Liu.) 
    
    S. Yan, H. Fang, N. O'Connor, and M. Liu are with the School of Electronic Engineering and SFI Insight Centre for Data Analytics at Dublin City University, Dublin 9, Ireland. (e-mail: sen.yan5@mail.dcu.ie; hongyuan.fang2@mail.dcu.ie; noel.oconnor@dcu.ie; mingming.liu@dcu.ie).
    
    T. Ward is with the School of Computing and SFI Insight Centre for Data Analytics at Dublin City University, Dublin 9, Ireland. (e-mail: tomas.ward@dcu.ie).
    
    J. Li is with the Department of Mechanical Engineering, University of Birmingham. Birmingham, B15 2TT, U.K. (e-mail: j.li.1@bham.ac.uk).}}    

    \maketitle
    \thispagestyle{empty}
    \pagestyle{empty}
    
    \begin{abstract}
        \label{sec:abstract}
        \input{data/abs}
    \end{abstract}


    \IEEEpeerreviewmaketitle

    
    \printacronyms[
        heading = section*, 
        name    = List of Abbreviations \& Acronyms, 
    ]
    \color{black}

    \section{Introduction}
        \label{sec: intro}
        \input{data/1_intro}

    \section{Literature Review}
        \label{sec: review}
        \input{data/2_review}

    \section{System Model}
        \label{sec: model}
        \input{data/3_model}

    \section{Research Methodology}
        \label{sec: method}
        \input{data/4_method}
    
    \section{Data \& Dataset}
        \label{sec: data}
        \input{data/5_data}

    \section{Experiment \& Results}
        \label{sec: exp}
        \input{data/6_exp}
    
    \section{Discussions}
        \label{sec: dis}
        \input{data/7_dis}
    
    \section{Conclusion \& Future Work}
        \label{sec: con}
        \input{data/8_con}

    \section*{Acknowledgement}
        \label{sec: ack}
        \input{data/ack}
        
    \bibliographystyle{IEEEtran}
    \bibliography{references}
    
    \input{data/bio}

\end{document}

%% file: data/abs.tex
Battery Electric Vehicles (BEVs) are increasingly significant in modern cities due to their potential to reduce air pollution. Precise and real-time estimation of energy consumption for them is imperative for effective itinerary planning and optimizing vehicle systems, which can reduce driving range anxiety and decrease energy costs. As public awareness of data privacy increases, adopting approaches that safeguard data privacy in the context of BEV energy consumption modeling is crucial. Federated Learning (FL) is a promising solution mitigating the risk of exposing sensitive information to third parties by allowing local data to remain on devices and only sharing model updates with a central server. Our work investigates the potential of using FL methods, such as FedAvg, and FedPer, to improve BEV energy consumption prediction while maintaining user privacy. We conducted experiments using data from 10 BEVs under simulated real-world driving conditions. Our results demonstrate that the FedAvg-LSTM model achieved a reduction of up to 67.84\% in the MAE value of the prediction results. Furthermore, we explored various real-world scenarios and discussed how FL methods can be employed in those cases. Our findings show that FL methods can effectively improve the performance of BEV energy consumption prediction while maintaining user privacy.

%% file: data/1_intro.tex
The transition from conventional internal combustion engine vehicles to \ac{ev} has emerged as a major global initiative due to increasing concerns related to the environment and public health. Governments worldwide are introducing policies, laws, and regulations to promote \ac{ev}s. For instance, the Members of the European Parliament recently approved regulations to encourage the production of zero- and low-emission vehicles, such as \ac{ev}s and plug-in hybrids\footnote{\url{https://www.europarl.europa.eu/news/en/press-room/20230210IPR74715/fit-for-55-zero-co2-emissions-for-new-cars-and-vans-in-2035}}. In Ireland, the Zero Emission Vehicles Ireland office has been established with a goal of switching an expected 30\% of the private car fleet to electric by 2030\footnote{\url{https://www.gov.ie/en/campaigns/18b95-zero-emission-vehicles-ireland/}}. Furthermore, numerous studies have indicated a significant reduction in air pollution, noise pollution, and dependence on fossil fuels \cite{Choma2020, mre2021, Xing2021} resulting from the adoption of \ac{ev}s.

Energy modeling is essential for policymakers to assess the impact on the grid, identify necessary infrastructure requirements based on vehicle-to-grid technology, and inform energy policy to support the transition to clean transportation to enhance the overall driving experience while mitigating transportation's environmental impact \cite{Hussain2020, Ding2022}. It is beneficial to automakers and companies in understanding the performance of \ac{ev}s and optimizing the energy usage patterns, leading to improved battery management and energy efficiency \cite{Madhusudhanan2020, Sagaria2021}. Besides, accurate predictions of energy consumption can enable drivers to make informed decisions regarding trip planning, route selection, and charging strategies \cite{Xiang2022, Liu2022}.

To this end, various methods have been applied in this research field, ranging from system dynamics models to statistical and data-driven models \cite{Zhang2020, Modi2020, Ullah2021, Ullah2021_2, Abdelaty2021, Chen2021_2, Mdziel2023}. However, with the increasing volume of data generated by connected vehicles, conventional energy modeling methods often lack the capacity to efficiently manage the produced data. Specifically, these methods often require data collection from various sources to create and train a centralized model. This centralized approach can introduce challenges in data management, such as data transfer and the risk of data leakage. Instead, \ac{fl} offers a promising solution that enables vehicles to share local models rather than raw data \cite{Wang2023} in a decentralized manner for better data privacy, security and scalability. While \ac{fl}-based methods may demand increased computational resources and more robust communication infrastructure \cite{Chen2021_4}, their superior learning capabilities can generally lead to improved model performance in accuracy, precision, and generalizability for many real-world applications \cite{Zahid2020, Zahid2020_1, Zahid2020_2, Bayat2021, Ibrahem2022, Firdaus2022, Barhoush2023}.

In our preliminary study \cite{Liu2021}, \ac{fl} methods show great potential to deal with privacy issues of connected vehicles. Edge-cloud computing is believed to be the key to meeting all challenges in big transportation data \cite{BaduMarfo2019, LpezAguilar2022, Habibzadeh2019, Hahn2021}. Based on these foundations, our research aims to achieve energy consumption modeling for \ac{ev}s based on \ac{fl} methods and investigates the effectiveness of models together with edge-cloud computing techniques. By evaluating and comparing the performance of different models for \ac{bev} energy consumption prediction with various configurations, we provide recommendations for their practical implementation in different real-world scenarios. The main contributions of our work are outlined below:

\begin{itemize}
    \item An extensive review of existing methods employed in previous research focusing on energy consumption modeling for \ac{ev}s.
    \item A comparative study is conducted on five local model candidates and five \ac{fl} algorithms to identify the optimal combination for \ac{bev} energy consumption modeling.
    \item The performances are evaluated for the models with different setups, including the number of iterations, data splitting ratio, and input data size.
    \item By leveraging distributed data sources and federated learning, the proposed method offers the potential to overcome data privacy concerns and enhance the accuracy and efficiency of \ac{ev} energy modeling.
    \item The proposed method is discussed in two centralized and decentralized edge-cloud computing structures for \ac{bev} energy consumption modeling. This enables more accurate predictions of \ac{ev} energy consumption in various real-world scenarios.
\end{itemize}

The structure of this paper is as follows. In \autoref{sec: review}, we provide a review of the relevant literature on \ac{ev} energy consumption modeling and \ac{fl} algorithms and their applications. The research question and object are explained in detail in \autoref{sec: model}. \autoref{sec: method} describes the methodology employed in our work, and \autoref{sec: data} provides the description and analysis of our dataset. \autoref{sec: exp} presents the experiment setups and corresponding results, and relevant discussions and analysis are included in \autoref{sec: dis}. Finally, we conclude our work in \autoref{sec: con} and discuss future plans and potential improvements.

%% file: data/2_review.tex
In this section, a comprehensive overview of the literature on energy consumption modeling for \ac{ev}s is provided at first. Secondly, an introduction to \ac{fl} algorithms is presented with some applications. The unique advantage of \ac{fl} algorithms in addressing privacy concerns is discussed at the end.

\subsection{Energy Consumption Modeling}

    The literature on estimating \ac{ev} energy consumption models has used three distinct categories of methods: white box, grey box, and black box. This section aims to provide a brief introduction to white and grey box methods while focusing primarily on black box methods.

    \subsubsection{White Box Methods}
        
        Methods that integrate physical constraints into their modeling process, known as white box methods, are invariably built upon theories of vehicle dynamics. These approaches typically estimate energy consumption by employing vehicle dynamics principles with data obtained from a single vehicle. For instance, \cite{Wu2015} explores power relationships with velocity, acceleration, and road grade for \ac{ev}s through statistical analysis and an analytical model based on vehicle dynamics principles. The data used in this study, including battery status, speed, acceleration, and position, was collected from a single test vehicle. For readers interested in further exploration of this topic, we recommend referring to the studies in \cite{Fiori2016, Ristiana2019, Miri2020, Zhao2021, Wankhede2022, Hull2023, Kolte2022}.

    \subsubsection{Grey Box Methods}

        Researchers have applied grey box methods to estimate \ac{ev} energy consumption by combining physical knowledge with data fitting techniques. These methods, capable of handling larger datasets through data fitting techniques, primarily focus on the input and output of models using data collected from individual vehicles. For example, the authors of \cite{Hong2016} investigated energy consumption modeling for \ac{ev}s using data collected from a custom-designed \ac{ev}. Their approach blends physics-based equations with empirical data, resulting in a hybrid power model that offers improved power consumption estimations and addresses range anxiety concerns. Furthermore, statistical techniques such as the Monte Carlo method \cite{Chen2021}, multinomial logistic regression \cite{Almaghrebi2020}, and Gaussian process regression \cite{Xu2020} have been utilized for similar purposes. For those interested in further exploring this topic, we recommend referring to the studies mentioned in \cite{Zhang2015, DeCauwer2015, Guo2020, Ji2022}.
        
    \subsubsection{Black Box Methods}
        
        Instead of physical knowledge, black box methods use data fitting techniques, such as \ac{ml} models, due to their capacity to make predictions and analyses of datasets. Some classic \ac{ml} models, such as Support Vector Machine \cite{Abdelaty2021}, Decision Tree and \ac{rf} \cite{Ullah2021}, have been applied and estimated in previous relevant work. Furthermore, deep learning methods such as \ac{ann} \cite{Ullah2021_2} and \ac{cnn} have also been employed to solve the same problem. Methods mentioned above are widely used to estimate \ac{ev} energy consumption based on clustered data collected from multiple vehicles. For instance, in \cite{Zhang2020}, the authors employed \ac{xgb} to forecast energy consumption based on the real-world data collected from 55 electric taxis from 2017 to 2018 in Beijing, China. The method demonstrated an impressive performance with a Root Mean Squared Error of 0.159 kWh. For further information, we refer interested readers to the explanations included in \cite{Zhang2020, Modi2020, Ullah2021, Ullah2021_2, Abdelaty2021, Chen2021_2, Mdziel2023}.

\subsection{Privacy-Aware in Transportation Data}

    The utilization of transportation data has gained significant traction in both research and industry applications. However, the integration of transportation data without due consideration for privacy can give rise to various risks and consequences. Recent scholarly investigations \cite{BaduMarfo2019, LpezAguilar2022, Habibzadeh2019, Hahn2021} have elucidated and examined the potential risks associated with the usage of big transportation data, including privacy infringement, data breaches, misuse, and the erosion of user trust. For instance, direct usage of source data may expose sensitive information. \cite{Hahn2021} highlighted how the divulgence of origin-destination information of individual users can compromise their privacy, allowing malevolent actors to deduce their residential or workplace locations. Another pertinent issue related to big transportation data exists in open datasets. As outlined in \cite{BaduMarfo2019}, by cross-referencing publicly available datasets from the New York City Taxi and Limousine Commission, the cash tips paid by celebrities could be discovered and released easily. The perception of data vulnerability and the potential for misuse or privacy breaches can undermine user trust, resulting in a hesitancy to share data or participate in related research endeavours.

    Simultaneously, the studies referenced above have also highlighted the advantages of incorporating privacy-aware methodologies in the context of transportation data. These benefits can be summarized as follows:

    \begin{itemize}
        \item From the users' perspective, the implementation of privacy-aware approaches is paramount to safeguarding user data, ensuring that sensitive information such as personal identities, location data, and behavioral patterns are adequately protected. By prioritizing privacy, users can have greater confidence in sharing their data and engaging in various services and platforms.
        \item From the operators' standpoint, adopting privacy-conscious practices is instrumental in building trust and cultivating a positive reputation among users. This, in turn, can foster increased user participation, collaboration, and support for initiatives that prioritize privacy protection. Operators who prioritize user privacy are more likely to attract and retain a loyal user base.
        \item In terms of privacy regulations and ethical considerations, adhering to privacy-aware practices is crucial for organizations to comply with relevant privacy laws and regulations. By implementing robust privacy measures, organizations demonstrate their commitment to respecting individuals' rights to privacy and autonomy. Aligning with ethical principles promotes a responsible and trustworthy environment for data collection and usage.
    \end{itemize}

\subsection{Federated Learning}
    To reduce the computational complexity and enhance the privacy preservation of the centralized approach, \ac{fl} was developed and recently it has received considerable attention due to its ability to enable multiple data holders, such as \ac{ev}s, to collaboratively train models without the need to share their underlying data, thereby providing a natural safeguard against data privacy breaches \cite{Mothukuri2021}. As a subfield of \ac{ml}, \ac{fl} methods bring revolutionary changes to the conventional system identification process. Instead of processing the data which is previously collected from a single vehicle or clustered from multiple vehicles, \ac{fl} methods enable each \ac{ev} to collect data and train the individual model locally without raw data sharing, so it is unnecessary to anonymize the data as what conventional methods did. Besides, the capability of \ac{fl} methods to deal with large-scale data make it possible to update the trained model and make the decision in time rather than optimize the system performance after another round of data clustering, processing and model training.

    In the seminal paper \cite{McMahan2016}, the \ac{sgd} and \ac{avg} algorithms were first proposed. A recent study \cite{Li2023} employed \ac{cnn} and bidirectional \ac{lstm} networks as local models and implemented \ac{avg} for global optimization. The study revealed that the performance of the model improved compared to the original local model. In another study \cite{Li2018}, the authors proposed the \ac{prox} framework, which aims to address the challenges posed by statistical and system heterogeneity in the context of \ac{fl}. 
    
    The domain of \ac{ev}s is characterized by significant variability in data distribution across different devices. Recent research has proposed novel approaches such as \ac{per} \cite{Arivazhagan2019} and \ac{rep} \cite{Collins2021} to address the challenges arising from this statistical heterogeneity. These methods introduce the concept of dividing the model into standardized layers, which perform weight sharing across the devices, and personalized layers, which are responsible for generating the final personalized output based on the local data of each device. These methods represent a promising strategy for mitigating the negative impacts of statistical heterogeneity in \ac{fl} for \ac{ev}s. Several survey papers, including \cite{Lim2020} and \cite{Alazab2022}, have listed various comparative experiments where different \ac{fl} approaches (e.g., \ac{avg} and \ac{per}) with diverse local models, such as \ac{lstm} and \ac{gru}, have been implemented and compared.

    Among the five \ac{fl} algorithms previously discussed, i.e., \ac{sgd}, \ac{avg}, \ac{prox}, \ac{per}, and \ac{rep}, each exhibits unique strengths within this domain. While some studies have compared specific pairs of these algorithms, a comprehensive comparison of all five is still warranted to offer a more thorough understanding of their respective advantages and limitations.

%% file: data/3_model.tex
In this section, we present our research object in terms of the research problem statement, the specification of vehicles, and the introduction of energy consumption calculation. The problem statement part provides a comprehensive overview of the research objectives, while the vehicle specification outlines the parameters of the vehicles used in the experiment. Additionally, the last part talks about the benchmark for calculating the energy consumed in transit.

\subsection{System Setup}
    
    We consider a scenario where a \ac{bev} company wishes to improve its customers' satisfaction and convenience in the aspect of battery capacity management by enhancing the performance of its energy consumption prediction model while respecting the customers' privacy (e.g., without sharing their trip detail data such as location, time, speed, or the status of the remaining charge of the vehicle battery). Thus, the aim of our research is to implement and improve the energy consumption prediction by keeping the collected private dataset locally but updating the \ac{ml} models globally with the help of different \ac{fl} methods. To do this, we assume that each \ac{bev} in the system is equipped with an onboard computing/communication unit, which is connected to our sensor group including various sensors capturing different trip attributes (e.g., velocity, trip distance, temperature and road slope) and is able to process the data collected, train a local machine-learning model accordingly and communicate and transfer data with other sections. 

    Similar to the model setup presented in our previous work \cite{Liu2021}, we now formulate the \ac{bev} energy consumption prediction as follows. Let $N$ be the number of \ac{bev}s in our system and $\mathbb{N}:=\{1,2,3, \dots, N\}$ be the set for indexing these \ac{bev}s. For a vehicle $i \in \mathbb{N}$, we assume the number of its trip records is $K$ and denote its $j^{th}$ trip by $\mathbb{T}_{i,j}$ where $j \in \{1, 2, \dots, K\}$. If our sensor group capture the trip features $M$ times in trip $\mathbb{T}_{i,j}$, the trip features will be divided into $M$ sets accordingly. Thus, at a given sampling moment $t \in \{1, 2, \dots, M\}$, the feature set is represented by $f_{i,j}^t$ and consequently, the trip $\mathbb{T}_{i,j}$ could be defined based on its feature sets as:
    
    \begin{equation} \label{equ: trip definition}
        \mathbb{T}_{i,j} = [f_{i,j}^1, f_{i,j}^2, \dots, f_{i,j}^M] 
    \end{equation}

    Assuming that the length of the trip feature set is $L$, we could denote the feature set at the sampling moment $t$ in the trip $\mathbb{T}_{i,j}$ by:

    \begin{equation} \label{equ: feature set definition}
        f_{i,j}^t = [f_{i,j}^t(1), f_{i,j}^t(2), \dots, f_{i,j}^t(L)] 
    \end{equation}

    \noindent where each $f_{i,j}^t(k)$ represents the $k^{th}$ feature in the feature set $f_{i,j}^t$, so accordingly, all the $K$ historical trip records of the vehicle $i$ could be defined as a collection of all trips during the day as:

    \begin{equation} \label{equ: history definition}
        \mathbb{D}^i = [\mathbb{T}_{i,1}, \mathbb{T}_{i,2}, \dots, \mathbb{T}_{i,K}] := [\mathbb{D}^i(1), \mathbb{D}^i(2), \dots, \mathbb{D}^i(L)] 
    \end{equation}

    \noindent where $\mathbb{D}^i(k)$ is a column vector of $\mathbb{D}^{i}$ where only the $k^{th}$ feature of the trip data is included. Assuming that the last feature, i.e., $\mathbb{D}^i(L)$, represents the energy consumption data $E^i$ of the vehicle $i$, we could use the other columns in $\mathbb{D}^{i}$ as the input feature $F^i$ for training the local model for the vehicle $i$, which could be defined as:
    
    \begin{equation} \label{equ: input definition}
        F^i := [\mathbb{D}^i(1), \mathbb{D}^i(2), \dots, \mathbb{D}^i(L-1)] 
    \end{equation}

    \noindent and output label data, i.e., the energy consumption data of the vehicle $i$ ($E^i$), could be defined as:
    
    \begin{equation} \label{equ: output definition}
        E^i := \mathbb{D}^i(L) 
    \end{equation}

    Finally, for a given time $t$, we define the past $m$ observations from the input and output sets as $F^i_{t-m+1:t}$ and $E^i_{t-m+1:t}$, respectively. The output vector has a dimension of $m$, representing the energy consumption, while the input matrix has a dimension of $m \times (L-1)$, representing the feature set at each given time point with the past $m$ steps in standard time units. Here, we consider the overall energy consumption data during the $m$ steps instead of the point-wise data of the vector, so we define the overall energy consumption within the past $m$ observations as:
    
    \begin{equation} \label{equ: overall energy consumption}
        \Hat{E}^i_{t-m+1:t} := \textbf{1}^T \cdot E^i_{t-m+1:t} \,\, \forall t \in \mathbb{K} 
    \end{equation}

    \noindent where $\textbf{1} \in \mathbb{R}^m$ is the column vector with all entries equal to 1, and $\mathbb{K} := \{m, m+1, \dots \}$ is a feasible indexing set. 

    Given the notation above, a local model training process for vehicle $i$ can be defined to find a local hypothesis function $\textbf{H}_i(\cdot)$ which is able to address the following problem:
    
    \begin{equation} \label{equ: problem statement}
        \left.
        \begin{alignedat}{-1}
            &\min_{\textbf{H}_i} \,\, \sum_{t \in \mathbb{K}}\left| \Hat{E}^i_{t-m+1:t} - \widetilde{E}^i_{t-m+1:t} \right|\\
            &\,\, \mathrm{s.t.} \,\, \widetilde{E}^i_{t-m+1:t} = \textbf{H}_i(F^i_{t-m+1:t}) \qquad
        \end{alignedat}
        \right.
    \end{equation}

\subsection{Research Object}

\begin{figure}[ht]
        \vspace{-0.1in}
        \centering
        \includegraphics[width=\linewidth]{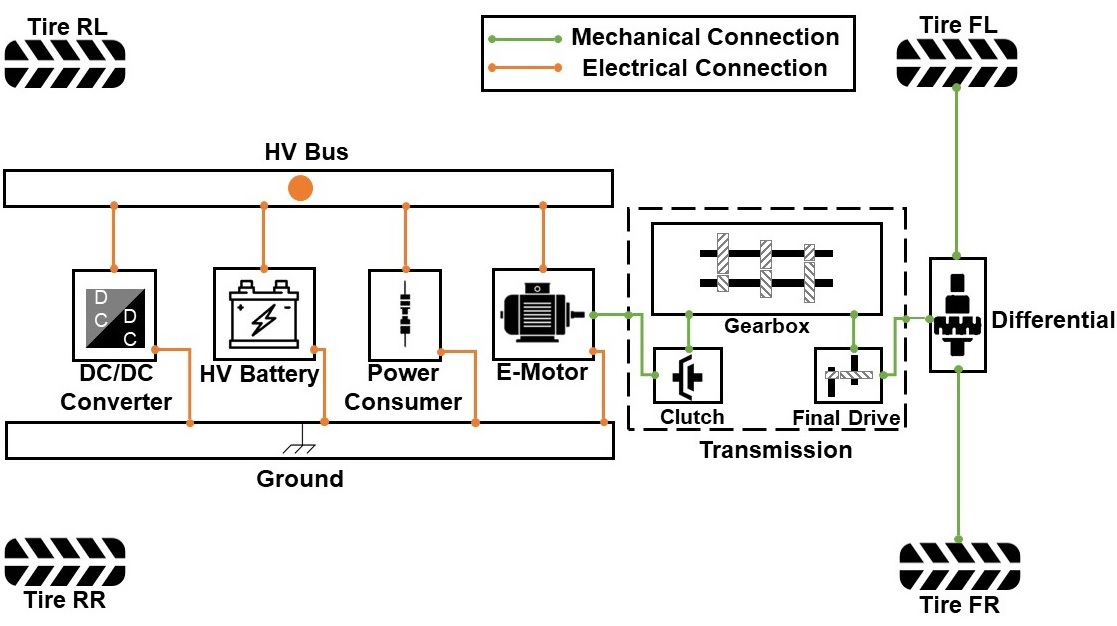}
        \caption{Vehicle architecture.}
        \label{fig: vehicle arch}
        \vspace{-0.1in}
    \end{figure}

    The studied vehicles are considered the same type of passenger \ac{bev}s as shown in \autoref{fig: vehicle arch}. It is equipped with an electric motor of a maximum of 102 kW at 3000 rpm and a battery pack with \textit{120s5p} cells for a total usable energy of 45 kWh and a maximum power of 160 kW. In order to maximize reliability, vehicle models are constructed through the industrial development software AVL Cruise M.

    In this model, battery management and electric motor control systems are involved to ensure that the vehicle works under safe conditions, i.e., the power limitation of the battery pack and a current limit for the e-motor to match the battery limitation. The most used functionalities of velocity-dependent regenerative braking, electrical-mechanic parallel braking and battery current limitation are also represented. By modifying the cockpit output signals, the electric drive control calculates the e-motor load signal in both traction and recuperation conditions. If the braking effect of the electric motor torque is below the driver’s request in terms of equivalent brake pressure, then the service brakes are supplied with pressure. 

\subsection{Energy Consumption Calculation}

    The model setup has been introduced in the previous sections, while here we present the method used to calculate energy consumption as the benchmark of our work based on the introduction in our previous work in \cite{Li2022_2}. The energy consumption of the studied \ac{ev} is all generated from its battery pack. In the model of the battery pack, battery cell current and voltage are applied in iterative calculations to simulate the battery cell dynamics. Starting from each iteration, the battery cell current $I_{bc}$ needs to be calculated firstly by the following formula:
    
    \begin{equation} \label{equ: battery current}
        I_{bc} = \frac {P_{bp}}{n_{bc} \cdot V_{bc}} \qquad 
    \end{equation}

    \noindent where $P_{bp}$ is the power of the battery pack, $n_{bc}$ is the number of battery cells and $V_{bc}$ is the terminal voltage of the battery cell. Here, a standard Resistor-Capacitor equivalent battery model is employed to expose the current-voltage dynamics of a lithium-ion battery cell. The battery’s voltage dynamics must obey:
    
    \begin{equation} \label{equ: battery voltage}
        \left\{
        \begin{alignedat}{-1}
            &V_{bc} = V_{oc}(SoC) - V_{p1} - R_0(SoC)I_{bc} \\
            &C_{p1}(SoC) \cdot \frac{\mathrm{d}V_{p1}}{\mathrm{d}t}= I_{bc}- \frac{V_{p1}}{R_{p1}(SoC)}\qquad 
        \end{alignedat}
        \right.
    \end{equation}

    \noindent where $V_{oc}$ is open circuit voltage, which is a function of the battery cell’s State of Charge $SoC$; $V_{p1}$ represents the transient voltage; $R_0$ and $R_{p1}$ indicate the effective series resistance and the transient resistance; and each of them is a function of the battery cell’s $SoC$; $C_{p1}$ indicates the transient capacity, which is a function of the $SoC$ as well. The $SoC$ of the battery cell is calculated by:
    
    \begin{equation} \label{equ: battery soc}
        SoC = SoC_0 - \int_{0}^{t} \frac{I_{bc}}{Q_{bc}}\,\mathrm{d}t\qquad 
    \end{equation}

    \noindent where $SoC_0$ is the initial $SoC$ of battery cells and $Q_{bc}$ is quantity of electric charge of battery cells. The battery cell data and calibrated model parameters are soured from AVL Cruise M. 
    
    This work simplifies the battery package computational complexity by not considering the unbalance between cells in the battery pack as their overall \ac{ev} range will not be affected. So far, the energy consumption of each cell $J$ used in the studied \ac{ev} can be calculated as:
    
    \begin{equation} \label{equ: cell energy consumption}
        J = \int_{0}^{t} (V_{oc}(SoC) \cdot I_{bc})\,\mathrm{d}t 
    \end{equation}

%% file: data/4_method.tex
In this section, we briefly introduce centralized and decentralized \ac{fl} structures, specify the \ac{ml} models used as our local model candidates and provide a comprehensive overview of some \ac{fl} optimization algorithms which have been widely employed in a variety of domains.

\subsection{Federated Learning Structures}

\begin{figure}[ht]
        \vspace{-0.1in}
        \centering
        \includegraphics[width=\linewidth]{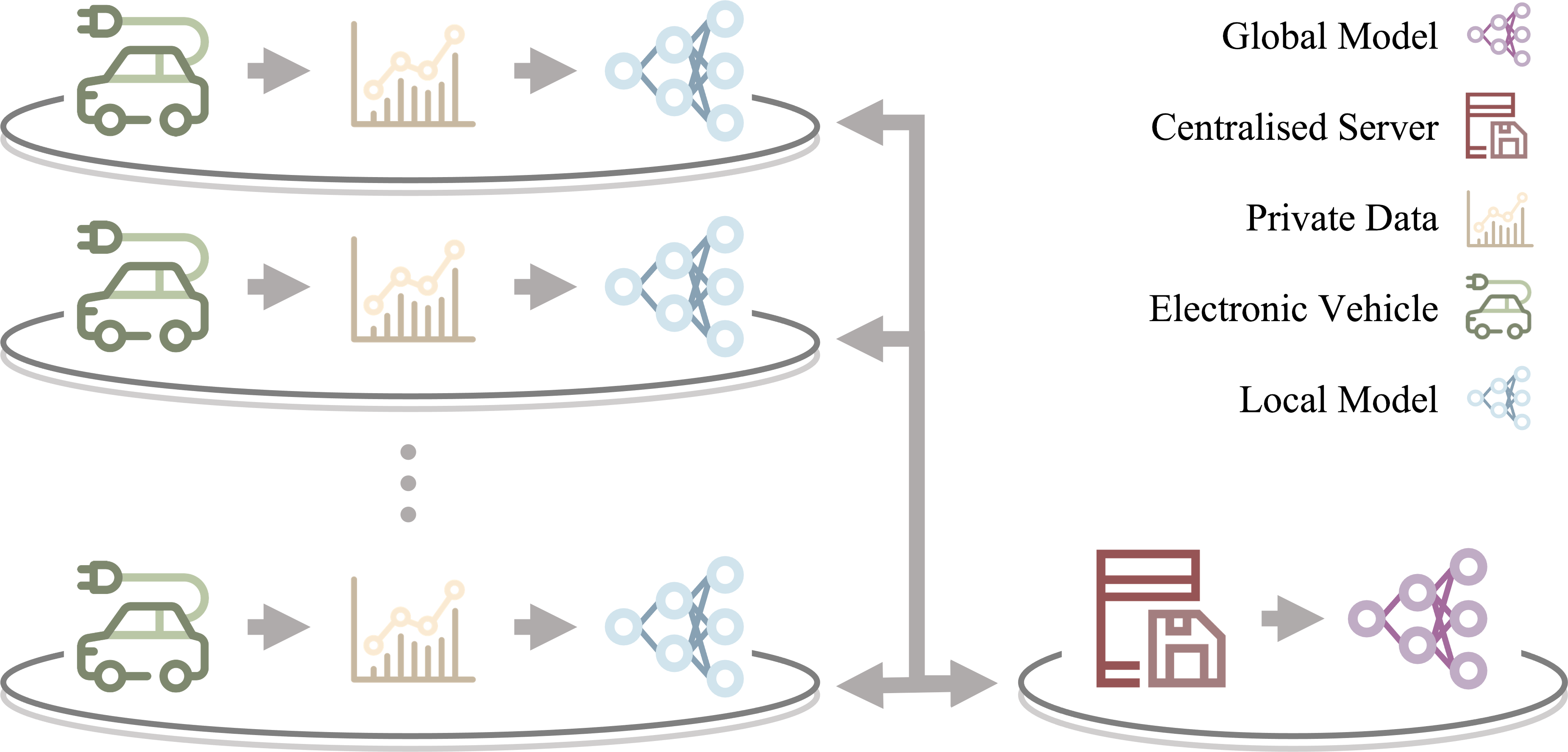}
        \caption{Centralized \ac{fl} architecture.}
        \label{fig: central}
        \vspace{-0.1in}
    \end{figure}

    In general, \ac{fl} methods can be implemented in both centralized and decentralized structures, which are tailored to various real-world conditions. Specifically, a centralized \ac{fl} implementation is illustrated in \autoref{fig: central}, where each local model is trained on private data collected from a \ac{bev} and transmits the requisite information to the centralized server. After calculating and updating the global model, the unique information is sent back to each \ac{bev} to update its local model. In this scenario, a powerful centralized server is indispensable, and the requirement for \ac{bev} computing units is relatively low.

\begin{figure}[ht]
        \vspace{-0.1in}
        \centering
        \includegraphics[width=\linewidth]{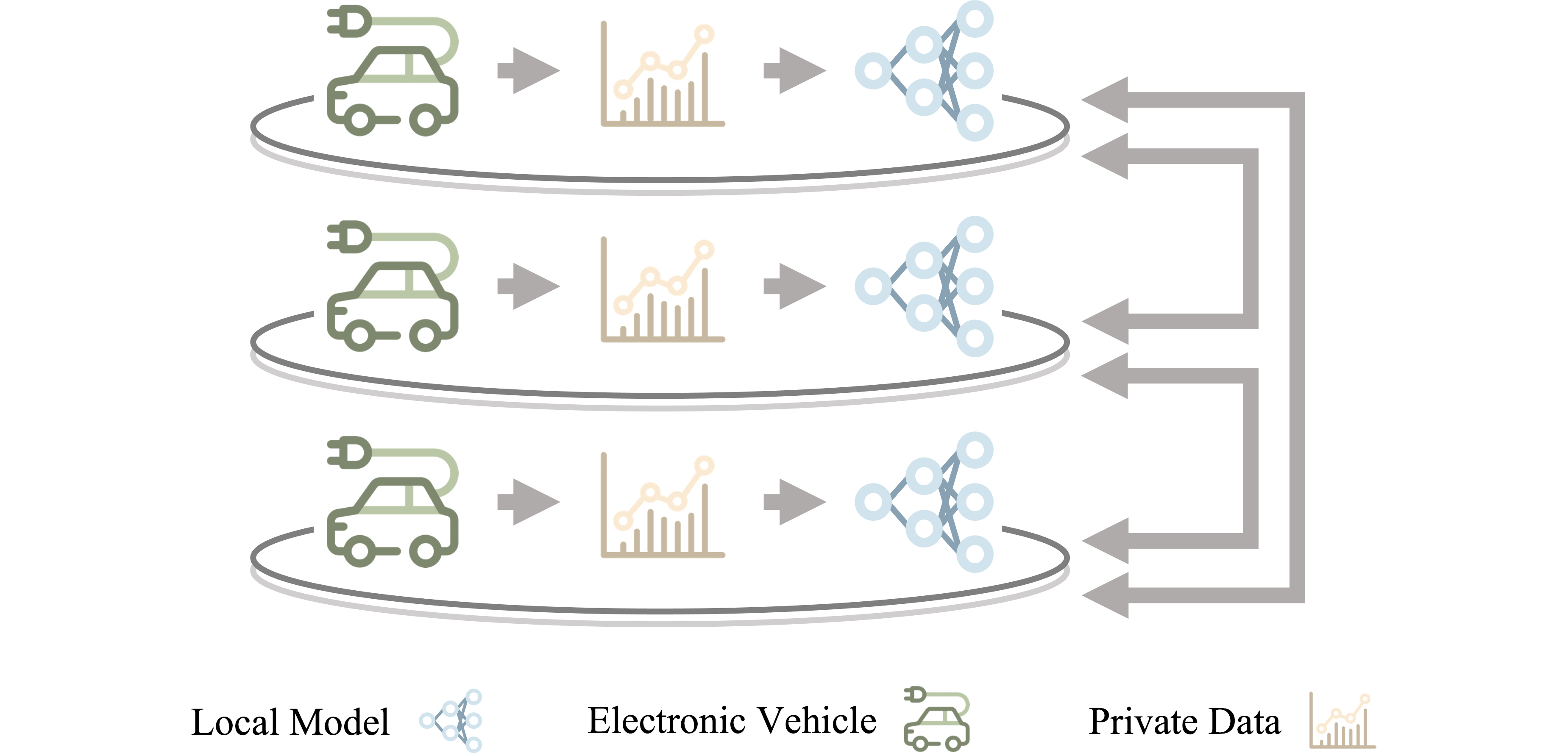}
        \caption{Decentralized \ac{fl} architecture.}
        \label{fig: decentral}
        \vspace{-0.1in}
    \end{figure}

    In contrast, for a decentralized \ac{fl} structure, as demonstrated in \autoref{fig: decentral}, \ac{bev}s share essential information and perform local model calculation and updating independently instead of relying on a server. In this case, the requirement for \ac{bev} computing units is greater, but the need for a centralized server is eliminated.

\subsection{Local Model Selection}

    Several basic \ac{ml} algorithms are adopted and compared to obtain the best candidate as our local model. As discussed in \autoref{sec: review}, we selected traditional \ac{ml} algorithms, e.g., \ac{rf} and \ac{xgb}, and deep learning algorithms, e.g., \ac{gru} and \ac{lstm}. The implementation of these models and their performance are provided in the following sections. 

\subsection{Federated Learning Algorithms}

    We apply different \ac{fl} optimization algorithms in our prediction model section and compare their results to find the best-performing algorithm as our basic design together with the local candidate model. As introduced in \autoref{sec: review}, \ac{sgd}, \ac{avg}, \ac{prox}, \ac{rep} and \ac{per} are widely used in similar systems, so these federated methods are adopted and their definitions are introduced in this part.

    \textit{\ac{sgd}:} the direct transposition of stochastic gradient descent to the federated setting. The gradients are averaged by the server proportionally to the number of training samples on each node and used to make a gradient descent step. The pseudocode for the \ac{sgd} algorithm is shown in \autoref{alg: FedSGD}.

    \begin{algorithm}
        \caption{\ac{sgd} for \ac{bev}s}
        \label{alg: FedSGD}
        \textbf{Notations:}
        \begin{algorithmic}
            \STATE The $V$ \ac{bev}s are indexed by $v$; the initial global gradient is $g_0$, the global gradient in the $t^{th}$ round is $g_t$ and the local gradient for vehicle $v$ in the $t^{th}$ round is $g_t^v$, the number of data points on vehicle $v$ is $n_v$, the total number of data points is $n$, the loss function is \textbf{Loss($\cdot$)}, and the local learning rate is $\eta$.
            \STATE
        \end{algorithmic}

        \textbf{Server executes:}
        \begin{algorithmic}
            \STATE initialize $g_0$
            \FOR{each round $t = 1, 2, \dots$}
                \FOR{each \ac{bev} $v \in V$ \textbf{in parallel}}
                    \STATE $g_{t+1}^v \gets$ SgdOpt$(v$, $g_t)$
                \ENDFOR
                \STATE $g_{t+1} \gets g_t - \eta \sum_{v=1}^{V} \frac {n_v}{n} \cdot g_{t+1}^v\qquad$
            \ENDFOR
            \STATE
        \end{algorithmic}

        \textbf{SgdOpt(}$v$, $g$\textbf{):}
        \begin{algorithmic}
            \STATE \textit{// run on vehicle} $v$
            \STATE $g \gets \triangledown$Loss$(g)$
            \RETURN $g$
        \end{algorithmic}
    \end{algorithm}

    \textit{\ac{avg}:} a generalization of \ac{sgd}, which enables local nodes to perform more than one batch update on local data and exchanges the updated weights rather than the gradients. The pseudocode for the \ac{avg} algorithm is shown in \autoref{alg: FedAvg and FedProx}.

    \textit{\ac{prox}:} a modified algorithm based on \ac{avg}, characterized by two modifications. Firstly, partial models that have not been fully trained are allowed, and all partial models participating in the training are integrated regardless of their accuracy. Secondly, the objective function of the local model is composed of the loss function plus the proximal term. The pseudocode for the \ac{prox} algorithm is shown in \autoref{alg: FedAvg and FedProx}.
    
    \begin{algorithm}
        \caption{\ac{avg} and \ac{prox} for \ac{bev}s}
        \label{alg: FedAvg and FedProx}
        \textbf{Notations:}
        \begin{algorithmic}
            \STATE The $V$ \ac{bev}s are indexed by $v$; the initial global weight is $w_0$, the global weight in the $t^{th}$ round is $w_t$ and the local weight for vehicle $v$ in the $t^{th}$ round is $w_t^v$, the number of data points on vehicle $v$ is $n_v$, the total number of data points is $n$, the loss function is \textbf{Loss($\cdot$)}, the local data of vehicle $v$ is $D_v$, the local minibatch size is $B$, the number of local epochs is $E$, and the local learning rate is $\eta$.
            \STATE
        \end{algorithmic}

        \textbf{Server executes:}
        \begin{algorithmic}
            \STATE initialize $w_0$
            \FOR{each round $t = 1, 2, \dots$}
                \STATE $V_t \gets$ (Full or partial set of $V$ \ac{bev}s)
                \FOR{each \ac{bev} $v \in V_t$ \textbf{in parallel}}
                    \STATE $w_{t+1}^v \gets$ AvgOpt$(v$, $w_t)$ \textbf{or} ProxOpt$(v$, $w_t$, $w_{t-1})$
                \ENDFOR
                \STATE $w_{t+1} \gets \sum_{v=1}^{V} \frac {n_v}{n} \cdot w_{t+1}^v\qquad$
            \ENDFOR
            \STATE
        \end{algorithmic}

        \textbf{AvgOpt(}$v$, $w$\textbf{):}
        \begin{algorithmic}
            \STATE \textit{// run on vehicle} $v$
            \STATE $\mathbb{B} \gets$ (split $D_v$ into batches of size $B$)
            \FOR{each local epoch $e$ from $1$ to $E$}
                \FOR{batch $b \in \mathbb{B}$}
                    \STATE $w \gets w - \eta \triangledown$Loss$(w;b)$
                \ENDFOR
            \ENDFOR
            \RETURN $w$\
            \STATE
        \end{algorithmic}

        \textbf{ProxOpt(}$v$, $w^t$, $w^{t-1}$\textbf{):}
        \begin{algorithmic}
            \STATE \textit{// run on vehicle} $v$\textit{; }$\mu$ \textit{is the proximal term coefficient} 
            \STATE $\mathbb{B} \gets$ (split $D_v$ into batches of size $B$)
            \FOR{each local epoch $e$ from $1$ to $E$}
                \FOR{batch $b \in \mathbb{B}$}
                    \STATE $w^t \gets w^t - \eta \triangledown($Loss$(w;b)+\frac{\mu}{2}\Vert w^t - w^{t-1}\Vert^2)$
                \ENDFOR
            \ENDFOR
            \RETURN $w_t$
        \end{algorithmic}
    \end{algorithm}
    
    \textit{\ac{per}:} a federated learning optimization method based on SGD which requires activating all clients and aggregating only the base layer parameters into individual personalized models. The pseudocode for the \ac{per} algorithm is shown in \autoref{alg: FedPer and FedRep}.

    \textit{\ac{rep}:} a federated learning optimization method based on SGD which requires activating all clients and focuses on learning a shared representation of the data, rather than learning individual models for each device. The pseudocode for the \ac{rep} algorithm is shown in \autoref{alg: FedPer and FedRep}.
    
    \begin{algorithm}
        \caption{\ac{per} and \ac{rep} for \ac{bev}s}
        \label{alg: FedPer and FedRep}
        \textbf{Notations:}
        \begin{algorithmic}
            \STATE The $V$ \ac{bev}s are indexed by $v$; the initial standardized layer weight matrix for vehicle $v$ is $W_S^{v, 0}$, the standardized layer weight matrix for vehicle $v$ in the $t^{th}$ round is $W_S^{v, t}$, the initial personalized layer weight matrix for vehicle $v$ is $W_P^{v, 0}$, the personalized layer weight matrix for vehicle $v$ in the $t^{th}$ round is $W_P^{v, t}$, the number of data points on vehicle $v$ is $n_v$, the total number of data points is $n$, the loss function is \textbf{Loss($\cdot$)}, and the local learning rate is $\eta$.
            \STATE
        \end{algorithmic}

        \textbf{Server executes:}
        \begin{algorithmic}
            \STATE \textit{// for \textbf{\ac{per}:} $W_S$ are base layers, $W_P$ are head layers;}
            \STATE \textit{// for \textbf{\ac{rep}:} $W_S$ are head layers, $W_P$ are base layers.}
            \STATE initialize $W_S^0 := \{W_S^{1, 0}, W_S^{2, 0}, \dots\}$
            \STATE initialize $W_P^{1, 0}, W_P^{2, 0}, \dots$
            \FOR{each round $t = 1, 2, \dots$}
                \FOR{each \ac{bev} $v \in V$ \textbf{in parallel}}
                    \STATE $(W_S^{v, t+1}, W_P^{v, t+1}) \gets$ Opt$(v$, $W_S^{v, t}$, $W_P^{v, t})$
                \ENDFOR
                \STATE $W_S^{t+1} \gets \sum_{v=1}^{V} \frac {n_v}{n} \cdot W_S^{v, t+1}\qquad$
            \ENDFOR
            \STATE
        \end{algorithmic}

        \textbf{Opt(}$v$, $W_S^v$ , $W_P^v$\textbf{):}
        \begin{algorithmic}
            \STATE \textit{// run on vehicle} $v$
            \STATE $(W_S^v$, $W_P^v) \gets (W_S^v$, $W_P^v) - \eta \triangledown$Loss$((W_S^v$, $W_P^v))$
            \RETURN $(W_S^v$, $W_P^v)$
        \end{algorithmic}
    \end{algorithm}

%% file: data/5_data.tex
This section presents the data in our dataset with both text descriptions and figure visualizations, followed by explanations of data pre-processing and feature selection in detail.

\subsection{Description \& Analysis}

    The dataset used in our work is generated based on \ac{ved}, a real-world dataset collected in Michigan, USA \cite{Oh2022}. We filtered 10 trips from \ac{ved} with a duration longer than 1,800 seconds. We then extracted their GPS coordinates and the altitude data as the input for our vehicle model introduced in \autoref{sec: model}, with the speed data set as the desired velocity of our vehicle model. The output, consisting of multiple trip attributes such as temperature, actual speed and trip distance, is then used as our dataset. In other words, our dataset is a collection of 10 tables of trip information for all 10 vehicles. Each table includes 1,800 rows and 12 columns, representing different attributes respectively. 

\begin{figure*}[ht]
        \vspace{-0.1in}
        \centering
        \includegraphics[width=\linewidth]{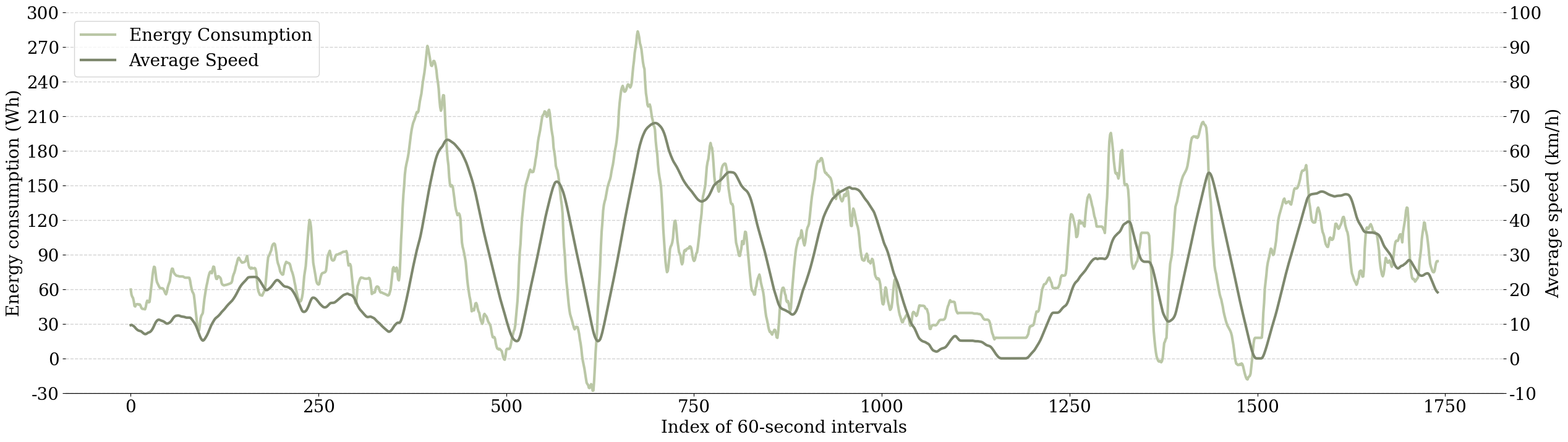}
        \caption{Effect of average speed on energy consumption in Vehicle 1.}
        \label{fig: speed and energy}
        \vspace{-0.1in}
    \end{figure*}

    We have explored the data for 10 vehicles and have discovered a correlation between their average speed and the total amount of energy consumed during a 60-second interval. For example, the relationship between average speed and energy consumption for Vehicle 1 is shown in \autoref{fig: speed and energy}. It is evident that there is a slight delay in the change of the average speed relative to the energy usage. Following our analysis, we have determined that this delay is approximately 23 seconds. 

\begin{figure*}[ht]
        \vspace{-0.1in}
        \centering
        \includegraphics[width=\linewidth]{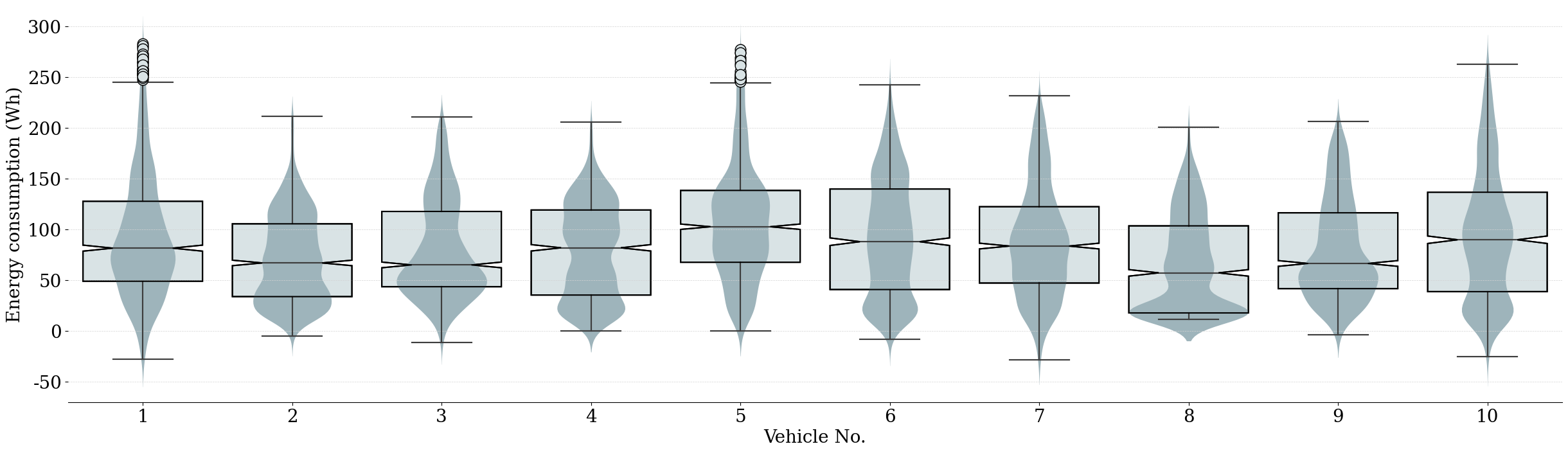}
        \caption{Box and violin plot of energy consumption for each vehicle.}
        \label{fig: boxplot}
        \vspace{-0.1in}
    \end{figure*}

    The energy consumption values in 60-second intervals have a wide range from a minimum of -28.15 Wh to a maximum of 283.38 Wh. The data analysis results revealed the distinct statistical features of individual vehicles, which suggests that they have been exposed to different road conditions and other circumstances in the 1,800 seconds of observation. Additionally, \autoref{fig: boxplot} indicates that there are a number of outliers in the data of Vehicle 1 and Vehicle 5, all of which are valid and cannot be disregarded. 

\subsection{Pre-processing}

    As introduced above, we found a strong connection between energy consumption and the average speed in a certain \ac{ts}, which makes it highly possible to predict energy consumption based on other features. Thus, we divide the data in each 60-second interval $I^i$ as follows:
    
    \begin{equation} \label{equ: features}
        I^i := \{F^i_{t-59}, F^i_{t-58}, \dots, F^i_{t-1}, F^i_{t}\} = F^i_{t-59:t} 
    \end{equation}

    \noindent where $F^i_{t}$ is the set of trip features at time $t$ for Vehicle $i$. With a full length of 1,800 seconds, the trip record data of one vehicle is divided into 1741 intervals.

    Moreover, standardization of the data is implemented to shift the distribution such that the mean is zero and the standard deviation is one. This preserves the essential information regarding the outliers, making the algorithm less susceptible to them, in comparison to Min-Max normalization.

\subsection{Feature Selection}

    As discussed in \autoref{sec: review}, most researchers working on similar research problems used speed, distance, acceleration and altitude as their features. Accordingly, we adopted these trip attributes and employed Principal Component Analysis and \ac{rf} Regression to select features generated by mathematical transformations (e.g., square, logarithm, and exponent), and based on the results, we selected acceleration $a$, speed $v$, the square root of speed $\sqrt{v}$, the cube of speed $v^3$ and the square root of the distance moved in each second $\sqrt{\Delta d}$ as the features for model training.

%% file: data/6_exp.tex
This section outlines our experimental cases, which encompass various experimental settings along with their corresponding results. By default, we assumed an input of the data from all vehicles would result in the best performance.

\subsection*{Case 1: Local Model Comparison}

    \textbf{Setups:} In this section, we applied various \ac{ml} models and compared their performance as the candidates of our local model deployed on each \ac{bev}. Here we implemented and evaluated \ac{rf}, \ac{xgb}, \ac{ann}, \ac{gru} and \ac{lstm} models.
        
    \textit{Machine Learning Models:} We initially applied the \ac{rf} and \ac{xgb} to implement the \ac{bev} energy consumption prediction on decentralized datasets, which were separately split into a training set and a test set with a ratio of $3:1$. In the setup of the \ac{rf} model, $16$ estimators were applied with a maximum tree depth of $9$ to run 16 jobs in parallel, while for the \ac{xgb} model, the regressor from the \ac{xgb} library with a gradient-boosted tree booster was used to run 50 jobs in parallel. 

    \textit{Deep Learning Models:} We adopt a similar structure for all the deep learning models in our work (i.e., \ac{ann}, \ac{gru} and \ac{lstm}) and replace the hidden layers with specific modules (i.e., dense layers, \ac{gru} layers and \ac{lstm} layers respectively). The deep learning model was set up based on three hidden layers containing 40, 32 and 16 neurons respectively with hyperbolic tangent as the activation function and the dropout layer was applied with a rate of $0.10$ and $0.20$ between the first and second and the second and third hidden layers separately. Coming from a dense layer, the model output of the model is a single value that by default captures the vehicle's overall energy consumption over the past 60 seconds. The loss function was chosen as the \ac{mae} which represents the average difference between the actual energy consumption of a \ac{bev} during a trip and the energy consumption predicted by the model. A lower \ac{mae} indicates that the model is better at predicting energy consumption and that the vehicle is likely to be more efficient, which can lead to reduced driving range anxiety and lower energy costs. The model was optimized by the Adam optimizer with the default learning rate, i.e., $1e^{-3}$. Each local dataset was split into batches with a size of $70$ with $65$ epochs. 

    \textbf{Results:} In \autoref{table: local}, a comparative analysis of the credible and competitive results of these models is presented. The value in each row of this table represents the average value of the \ac{mae} for the model trained on the corresponding vehicle's training set and tested on all vehicles' test sets. 

    \begin{table}[ht]
        \caption{Local model performance (\ac{mae}) on each vehicle (Wh).}
        \label{table: local}
        \begin{tabularx}{\linewidth}{@{\extracolsep{\fill}}c c c c c c}
            \toprule
            ID  & \ac{rf}      & \ac{xgb}     & \ac{ann}     & \ac{gru}     & \ac{lstm}    \\
            \midrule
            V1  & 9.5375  & 9.0951  & 6.9207  & 5.3355  & \cellcolor[RGB]{204, 239, 252}5.1759  \\
            V2  & 10.4122 & 9.7776  & 8.3997  & 6.8031  & \cellcolor[RGB]{204, 239, 252}6.7997  \\
            V3  & 10.1580 & 9.4953  & 7.2892  & 6.3033  & \cellcolor[RGB]{204, 239, 252}4.4117  \\
            V4  & 8.9370  & 8.6519  & 7.8394  & 6.6888  & \cellcolor[RGB]{204, 239, 252}4.6765  \\
            V5  & 10.7788 & 10.6958 & 8.6680  & \cellcolor[RGB]{204, 239, 252}5.6422  & 6.1466  \\
            V6  & 9.2424  & 8.8490  & 8.2999  & \cellcolor[RGB]{204, 239, 252}4.9306  & 5.8707  \\
            V7  & 11.5519 & 10.5905 & 7.0247  & 5.3347  & \cellcolor[RGB]{204, 239, 252}4.9729  \\
            V8  & 8.9563  & 9.0859  & 9.0698  & 7.5178  & \cellcolor[RGB]{204, 239, 252}5.7530  \\
            V9  & 12.1133 & 11.5608 & 11.0439 & 10.7237 & \cellcolor[RGB]{204, 239, 252}8.6670  \\
            V10 & 9.0282  & 8.8239  & 11.7233 & \cellcolor[RGB]{204, 239, 252}7.1755  & 8.1254  \\
            \bottomrule
        \end{tabularx}
    \end{table}

    \textbf{Evaluations:} As observed in \autoref{table: local}, the \ac{lstm} model demonstrates the best performance in terms of \ac{mae}. It is noteworthy, however, that different datasets with varying attributes may result in different best-performing models. As such, in our study, we selected the \ac{lstm} model as the local model for our \ac{fl} framework. Nonetheless, we acknowledge the need for adjusting the selection of the local model based on the prevailing conditions and characteristics of the dataset in future applications.
    
\subsection*{Case 2: Federated Learning Algorithm Comparison}

    \textbf{Setups:} As reviewed in \autoref{sec: method}, five algorithms, i.e., \ac{sgd}, \ac{avg}, \ac{prox}, \ac{per} and \ac{rep}, were implemented and evaluated at this stage. It is evident that these algorithms share a similar structure and basic logic, with the exception of \ac{sgd}, which calculates the gradient rather than weights, and \ac{per} and \ac{rep}, which employ an additional parameter, namely the number of personalized layers. The performances of the different local model candidates, as shown in \autoref{table: local}, led to the selection of \ac{lstm} due to its lowest \ac{mae} scores. Given the low complexity of the model, only one personalized \ac{lstm} unit was set, with the other layers being standardized. 

    \textbf{Results:} The comparative analysis of different \ac{fl} algorithms is presented in \autoref{table: global}. The value in each row of this table represents the average value of the \ac{mae} for the model trained on the corresponding vehicle's training set and tested on all vehicles' test sets. 

    \begin{table}[ht]
        \caption{\ac{fl} method performance (\ac{mae}) on each vehicle (Wh).}
        \label{table: global}
        \begin{tabularx}{\linewidth}{@{\extracolsep{\fill}}c c c c c c}
            \toprule
            ID  & \ac{sgd} & \ac{avg} & \ac{prox}& \ac{per} & \ac{rep}  \\
            \midrule
            V1  & 5.4300 & 4.0245 & 4.9574 & \cellcolor[RGB]{204, 239, 252}3.8197 & 4.4066  \\
            V2  & 6.3252 & 4.8791 & 6.8512 & \cellcolor[RGB]{204, 239, 252}4.8160 & 5.6075  \\
            V3  & 4.3714 & 3.8117 & 4.4467 & \cellcolor[RGB]{204, 239, 252}3.6836 & 4.1196  \\
            V4  & 4.2921 & 3.4877 & 4.4171 & \cellcolor[RGB]{204, 239, 252}3.3153 & 3.9459  \\
            V5  & 4.9364 & \cellcolor[RGB]{204, 239, 252}4.5298 & 6.4715 & 4.6095 & 5.5250  \\
            V6  & 5.1704 & \cellcolor[RGB]{204, 239, 252}3.7653 & 5.5837 & 3.8582 & 4.3317  \\
            V7  & 4.5819 & \cellcolor[RGB]{204, 239, 252}4.1943 & 5.2512 & 4.3598 & 4.9148  \\
            V8  & 5.1873 & 4.3449 & 6.0334 & \cellcolor[RGB]{204, 239, 252}4.1380 & 5.1330  \\
            V9  & 7.5800 & \cellcolor[RGB]{204, 239, 252}7.4539 & 9.9346 & 7.5819 & 9.5547  \\
            V10 & 8.9892 & \cellcolor[RGB]{204, 239, 252}7.5734 & 7.7679 & 7.6032 & 7.5043  \\
            \bottomrule
        \end{tabularx}
    \end{table}

    \textbf{Evaluations:} As observed in \autoref{table: global}, \ac{avg} and \ac{per} exhibit comparable performances on the dataset utilized in the study. As discussed in \autoref{sec: review}, the efficacy of personalized \ac{fl} approaches is highly reliant on the personalized layers selected. Hence, the reason for \ac{per} and \ac{rep} failing to surpass the performance of \ac{avg} could be attributed to the relatively low complexity of the local models employed in the study. Notwithstanding the comparable performances of \ac{avg} and \ac{per}, the time cost associated with both methods must be carefully considered. Taking into account the time required to complete 15 \ac{itr} of \ac{avg} and \ac{per}, i.e., around 49 and 58 minutes respectively, we ultimately decided to adopt \ac{avg} as the global \ac{fl} method for the subsequent experiments.
    
\subsection*{Case 3: Impact of Iteration}

    \textbf{Setups:} To investigate the impact of iteration numbers, the performances of the models with \ac{itr} of 15, 30, 45 and 60 were examined based on the \ac{lstm}-\ac{avg} structure. These results were compared to the performance of the local models (i.e., with 0 \ac{itr}), which served as the baselines.

    \textbf{Results:} \autoref{table: itr} presents the impact of iteration on the model performance. The value in each row of this table represents the \ac{mae} value of the model trained and tested on the corresponding vehicle's training and test sets.

    \begin{table}[ht]
        \caption{\ac{itr} on model performance (Wh).}
        \label{table: itr}
        \begin{tabularx}{\linewidth}{@{\extracolsep{\fill}}c c c c c c}
            \toprule
            ID  & \ac{lstm}   & 15 \ac{itr} & 30 \ac{itr} & 45 \ac{itr} & 60 \ac{itr}  \\
            \midrule
            V1  & 2.0949 & 1.0118 & \cellcolor[RGB]{204, 239, 252}0.6384 & 0.7127 & 0.6771  \\
            V2  & 1.4466 & 0.7181 & 0.5268 & \cellcolor[RGB]{204, 239, 252}0.3533 & 0.3961  \\
            V3  & 1.5723 & 0.5057 & 0.5268 & 0.5463 & \cellcolor[RGB]{204, 239, 252}0.3087  \\
            V4  & 1.6571 & 0.8222 & 0.8466 & \cellcolor[RGB]{204, 239, 252}0.4838 & 0.6151  \\
            V5  & 1.1317 & 0.5640 & 0.5039 & 0.5823 & \cellcolor[RGB]{204, 239, 252}0.2765  \\
            V6  & 2.2842 & 0.9905 & 0.6548 & \cellcolor[RGB]{204, 239, 252}0.5882 & 0.8435  \\
            V7  & 1.4967 & 0.7576 & 0.5935 & 0.4758 & \cellcolor[RGB]{204, 239, 252}0.4244  \\
            V8  & 1.7994 & 0.8203 & 0.8849 & 0.6794 & \cellcolor[RGB]{204, 239, 252}0.4169  \\
            V9  & 1.3155 & 0.6690 & 0.5529 & 0.3905 & \cellcolor[RGB]{204, 239, 252}0.2754  \\
            V10 & 2.2864 & 0.9877 & 0.7184 & \cellcolor[RGB]{204, 239, 252}0.5738 & 0.7174  \\
            \bottomrule
        \end{tabularx}
    \end{table}

    \textbf{Evaluations:} As demonstrated in \autoref{table: itr}, an increase in the number of \ac{itr} is accompanied by a decrease in the \ac{mae} for half of the objects (Vehicles 3, 5, 7, 8, and 9), indicating an improved model performance. However, for four objects (Vehicles 2, 4, 6, and 10), the best model performance occurred after 45 \ac{itr}. Vehicle 1 had the lowest \ac{mae} after 30 \ac{itr}. This trend can be attributed to the unique nature of the local data. Furthermore, while the model performance improved for some vehicles, the training process's time requirements increased. Specifically, executing 15, 30, 45, and 60 \ac{itr} required approximately 49, 98, 148, and 197 minutes, respectively, which implies that each iteration took about 49 minutes. After careful consideration of both model performance and time requirements, we chose to execute 15 \ac{itr} for this study.
    
\subsection*{Case 4: Impact of Data Split Ratio}

    \textbf{Setups:} Based on the \ac{lstm}-\ac{avg} structure, we conducted a comparative analysis of model performance across varying train-validation-test splitting ratios, i.e., 4:1:5, 5:1:4, 6:1:3, 7:1:2, and 8:1:1 to investigate the impact of data split ratio.

    \textbf{Results:} The impact of data-splitting ratios on model performance is reported in \autoref{table: split}. The value in each row of this table represents the \ac{mae} value of the model trained and tested on the corresponding vehicle's training and test sets.

    \begin{table}[ht]
        \caption{Impact of splitting ratio on model performance (Wh).}
        \label{table: split}
        \begin{tabularx}{\linewidth}{@{\extracolsep{\fill}}c c c c c c}
            \toprule
            ID  & 4:1:5  & 5:1:4  & 6:1:3  & 7:1:2  & 8:1:1   \\
            \midrule
            V1  & 1.3546 & 1.3068 & 1.0626 & \cellcolor[RGB]{204, 239, 252}0.7437 & 1.0118  \\
            V2  & 0.7424 & 0.6625 & 0.8641 & \cellcolor[RGB]{204, 239, 252}0.6374 & 0.7181  \\
            V3  & 1.0652 & 0.9029 & 0.7644 & 0.7448 & \cellcolor[RGB]{204, 239, 252}0.5057  \\
            V4  & 1.2096 & 1.0056 & 1.0278 & \cellcolor[RGB]{204, 239, 252}0.7597 & 0.8222  \\
            V5  & 0.9935 & 0.7111 & 0.6575 & 0.7352 & \cellcolor[RGB]{204, 239, 252}0.5640  \\
            V6  & 1.2573 & 1.4778 & 1.1662 & 1.1626 & \cellcolor[RGB]{204, 239, 252}0.9905  \\
            V7  & 1.0656 & 0.9155 & 0.6053 & \cellcolor[RGB]{204, 239, 252}0.5805 & 0.7576  \\
            V8  & 1.2109 & 1.0480 & 1.0135 & 0.9522 & \cellcolor[RGB]{204, 239, 252}0.8203  \\
            V9  & 0.7191 & 1.1471 & 0.9642 & 0.7911 & \cellcolor[RGB]{204, 239, 252}0.6690  \\
            V10 & 1.6850 & 1.2798 & 1.6003 & 1.0928 & \cellcolor[RGB]{204, 239, 252}0.9877  \\
            \bottomrule
        \end{tabularx}
    \end{table}

    \textbf{Evaluations:} Based on the results presented in the table, it can be observed that a data splitting ratio of 7:1:2 or 8:1:1 tends to yield the best performance on our dataset compared to the other ratios. In particular, a ratio of 8:1:1 demonstrates superior performance across all metrics. This indicates that an increase in the quantity of training data facilitates enhanced learning from the input under our experimental conditions. Consequently, in this work, we have selected a data splitting ratio of 8:1:1 due to its superior overall performance.
    
\subsection*{Case 5: Impact of Input Data Size}

    \textbf{Setups:} Concerning the input data sizes, we utilized data inputs with varying \ac{ts} of 60, 90, 120, 150, and 180 timestamps to train the model based on \ac{lstm}-\ac{avg}, with the subsequent performance evaluation.

    \textbf{Results:} The results of the experiments are presented in \autoref{table: timestamp}. The value in each row of this table represents the \ac{mae} value of the model trained and tested on the corresponding vehicle's training and test sets.

    \begin{table}[ht]
        \caption{Impact of input data size (\ac{ts}) on model performance (Wh).}
        \label{table: timestamp}
        \begin{tabularx}{\linewidth}{@{\extracolsep{\fill}}c c c c c c}
            \toprule
            ID  & 60 \ac{ts}   & 90 \ac{ts}  & 120 \ac{ts}  & 150 \ac{ts}  & 180 \ac{ts}   \\
            \midrule
            V1  & \cellcolor[RGB]{204, 239, 252}1.0118 & 2.0234 & 2.4310 & 3.4794 & 7.2946  \\
            V2  & \cellcolor[RGB]{204, 239, 252}0.7181 & 1.0923 & 1.2904 & 1.7201 & 3.9600  \\
            V3  & \cellcolor[RGB]{204, 239, 252}0.5057 & 1.5067 & 1.6543 & 2.4812 & 3.4214  \\
            V4  & \cellcolor[RGB]{204, 239, 252}0.8222 & 1.5299 & 2.0712 & 2.9217 & 3.2603  \\
            V5  & \cellcolor[RGB]{204, 239, 252}0.5640 & 1.1884 & 1.4450 & 2.1414 & 2.4692  \\
            V6  & \cellcolor[RGB]{204, 239, 252}0.9905 & 1.5533 & 1.7502 & 2.6796 & 3.7042  \\
            V7  & \cellcolor[RGB]{204, 239, 252}0.7576 & 1.1970 & 1.6349 & 1.7855 & 2.5312  \\
            V8  & \cellcolor[RGB]{204, 239, 252}0.8203 & 1.6027 & 1.7615 & 2.6084 & 3.5714  \\
            V9  & \cellcolor[RGB]{204, 239, 252}0.6690 & 0.9991 & 0.9454 & 1.5623 & 1.6180  \\
            V10 & \cellcolor[RGB]{204, 239, 252}0.9877 & 2.0208 & 2.2899 & 3.8614 & 8.7896  \\
            \bottomrule
        \end{tabularx}
    \end{table}

    \textbf{Evaluations:} As can be seen in the table, the increase in input data size is typically associated with an increase in \ac{mae}, indicating poorer model performance. This result may be attributed to the decrease in the number of windows due to the increase in window size, resulting in a reduction of training data and consequent degradation of model performance.
    
\subsection*{Case 6: Decentralized Approaches}

    \textbf{Setups:} In order to incorporate various real-world scenarios, experiments were conducted based on the chosen \ac{fl} approach in a decentralized setup. Using results from \ac{lstm}-based local models, we selected Vehicles 3, 4, and 6 as the top three performers (G), and Vehicles 2, 9, and 10 as the three weakest performers (W) to investigate the performance of decentralized \ac{fl} approaches. From these 6 models, different test groups were formed based on the number of weaker performers as below:

    \begin{itemize}
        \item {Three weak performers (0G+3W)}
        \item {One good performer with two weak performers (1G+2W)}
        \item {Two good performers with one weak performer (2G+1W)}
        \item {Three good performers (3G+0W)}
    \end{itemize}

    \textbf{Results:} The findings are presented in \autoref{table: dec}.

    \begin{table}[ht]
        \caption{Performance (\ac{mae}) of decentralized \ac{avg} method (Wh).}
        \label{table: dec}
        \begin{tabularx}{\linewidth}{@{\extracolsep{\fill}}c c c c c c}
            \toprule
            ID  & \ac{lstm}   & 0G+3W  & 1G+2W  & 2G+1W  & 3G+0W   \\
            \midrule
            V1 (W) & 2.0949 & 0.2403 & -      & -      & -       \\
            V2 (G) & 1.4466 & -      & -      & -      & 0.1956  \\
            V5 (G) & 1.1317 & -      & 0.1929 & 0.1892 & 0.2687  \\
            V6 (W) & 2.2842 & 0.3488 & 0.2770 & -      & -       \\
            V9 (G) & 1.3155 & -      & -      & 0.1413 & 0.2301  \\
            V10 (W)& 2.2864 & 0.2815 & 0.3700 & 0.3106 & -       \\
            \bottomrule
        \end{tabularx}
    \end{table}    

    \textbf{Evaluations:} The results indicate that after 15 \ac{itr} using the \ac{avg} algorithm, the performance of all local models in each case improved, demonstrating the efficacy of decentralized \ac{fl} methods in enhancing \ac{bev} energy consumption modeling. However, an increase in the number of good performers does not always equate to an improvement in performance. Specifically, the \ac{mae} value of Vehicle 10 increased when interacting with Vehicles 5 and 6 (1G+2W) compared to being aggregated with two weak performers (Vehicles 1 and 6). Conversely, the interaction among good performers does not always result in the most suitable setup for an individual vehicle model. For example, the best model performance for Vehicle 2 occurs when interacting with Vehicles 9 and 10 (2G+1W), but it becomes worse when aggregated with Vehicles 5 and 9 (3G+0W).

%% file: data/7_dis.tex
This section presents a comprehensive discussion of the experiment results in terms of the performance of a decentralized structure and \ac{fl} applications in the real world.

\subsection{Decentralized Aggregation}

    In the experiments presented in Section \ref{sec: exp}, it was observed that the aggregation results of models trained on separate private data were not directly affected by the number or percentage of good performers, but rather a suitable group for aggregation is more crucial for individual models. For instance, for two good performers, Vehicle 5 and 9, the results from interacting with Vehicle 10 were significantly better than those with Vehicle 2. This suggests that the similarity and dissimilarity of data attributes are crucial factors, and exploring the similarity of user behaviours and driving patterns is necessary for achieving better aggregation results.

\subsection{Real-World Application}

\begin{figure}[ht]
        \vspace{-0.1in}
        \centering
        \includegraphics[width=\linewidth]{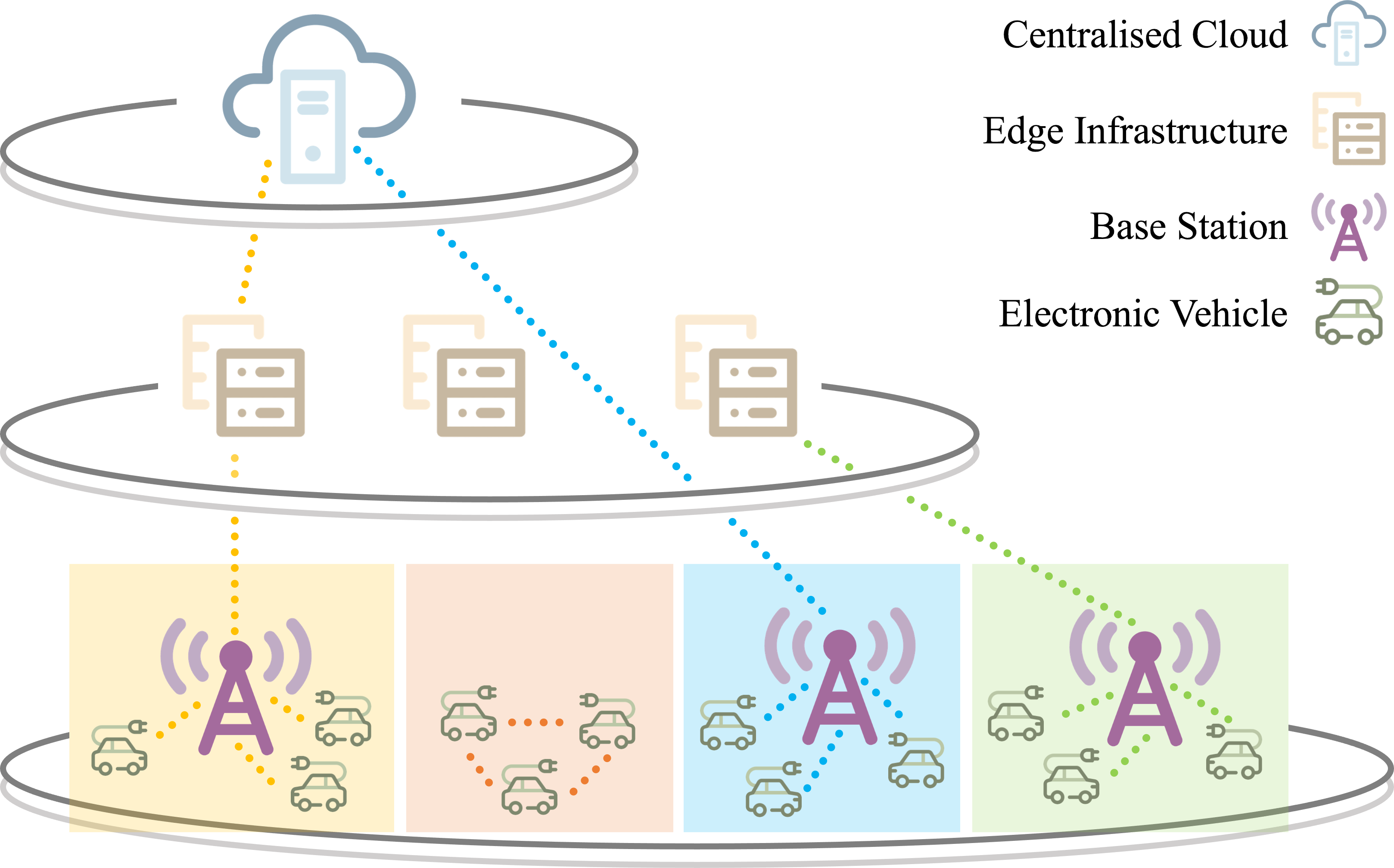}
        \caption{System framework based on edge computing.}
        \label{fig: framework}
        \vspace{-0.1in}
    \end{figure}

    \ac{fl} methods are well-suited to edge-cloud computing frameworks. We propose the application of our work in a real-world edge-computing-based system, the framework of which is illustrated in \autoref{fig: framework}. The top layer comprises a cloud data center, the middle layer consists of multiple edge infrastructures and the bottom layer is composed of base stations and \ac{bev}s. Different methods of data transmission and calculation have been considered and marked with different colours to accommodate various communication situations and hardware conditions. We now provide details of the four blocks from left to right in this framework:

    \begin{itemize}
        \item {The first block in yellow illustrates the classic centralized method, wherein data provided by \ac{bev}s is transferred to the edge server through the base station and then passed to and calculated by the cloud server. This approach does not necessitate \ac{bev}s to calculate, but they can request the edge server to calculate the weights in local models.}
        \item {The subsequent block in red depicts the fully decentralized method, wherein \ac{bev}s communicate with each other directly. This necessitates that the data computing and transmission capabilities of \ac{bev}s be highly robust and reliable.}
        \item {The third block in blue represents cloud computing, wherein data provided by \ac{bev}s is sent to the cloud server directly through the base station, without the requirement of edge infrastructures.}
        \item {Lastly, the block in green indicates that data is only transmitted to the edge server, thus requiring the edge server to possess a high capability to calculate and deploy the data.}
    \end{itemize}

%% file: data/8_con.tex
This paper aims to establish a privacy-aware energy consumption modeling framework for connected \ac{bev}S. To achieve that, we investigated the performance of various \ac{fl} algorithms in \ac{bev} energy consumption modeling. Based on comparative experimental results and relevant discussions in the previous section, we initially identified the superiority of \ac{lstm} and \ac{avg} as the local model and \ac{fl} algorithm for our dataset. Subsequently, we examined the impact of \ac{itr}, data splitting ratios, and input data sizes on model performance. Through the linear correlation between the number of \ac{itr} and time spent, and considering the corresponding model performances, we recommend the application of 15 \ac{itr} in our task. Regarding the model performance based on different data splitting ratios, we found that 8:1:1 is a suitable choice for datasets similar to ours. Furthermore, we observed a decrease in model performance corresponding to the increase in input data size, indicating that a \ac{ts} of 60 timestamps leads to relatively good results. After conducting our experiments, we observed a significant reduction in the \ac{mae} of our prediction results. Specifically, we achieved a reduction of up to 67.84\% (from 1.5723 to 0.5057 for Vehicle 3) by performing 15 \ac{itr}. The average \ac{mae} value for all vehicles was around 0.7847, indicating that the mean error for predicting energy consumption within one minute is approximately 0.8 Wh.

In addition, we explored decentralized \ac{fl} approaches for \ac{bev} energy consumption modeling by testing the selected \ac{fl} framework on the three best and three worst performers. The results demonstrated the effectiveness of \ac{fl} methods in this prediction task. Moreover, we provided a detailed description of how to apply \ac{fl} methods with an edge-cloud computing framework with various setups. We believe that the explorations and relevant analysis we conducted are meaningful and beneficial for future researchers, business operators, and policymakers.

Notwithstanding our extensive experiments and analyses, there are still some potential avenues for improvement. Firstly, the scope of our study could be extended by including a wider range of local model candidates and \ac{fl} methods. Additionally, in this paper, we made certain assumptions to simplify the presence of cell heterogeneity, which could be addressed by absorbing and analyzing extensive datasets in our future work. Furthermore, augmenting the size of our dataset could help us design better models that enable us to explore personalized \ac{fl} further. Besides, investigating the performance of models trained on simulated data on a real-world dataset is another aspect that can be explored in future work. Addressing these limitations would contribute to a more comprehensive understanding of \ac{fl} methods for \ac{bev} energy consumption modeling.

%% file: data/ack.tex
The authors would like to thank Rajesh Dande and Samaksh Chandra from Dublin City University for their help with the experiments, and also AVL List GmbH (AVL) for providing access to its advanced simulation technologies and software technical support within the frame of the University Partnership Program. This research was conducted with the financial support of Science Foundation Ireland \textit{21/FFP-P/10266} and \textit{12/RC/2289\_P2} at Insight the SFI Research Centre for Data Analytics at Dublin City University. 

The VED dataset is available at \url{https://github.com/gsoh/VED}. Our processed data used for this study is available on GitHub (URL: \url{https://github.com/SFIEssential/FedBEV}).

%% file: data/bio.tex
\vspace{-0.5\baselineskip}

\begin{IEEEbiography}[{
    \includegraphics[width=1in,height=1.33in,clip,keepaspectratio]{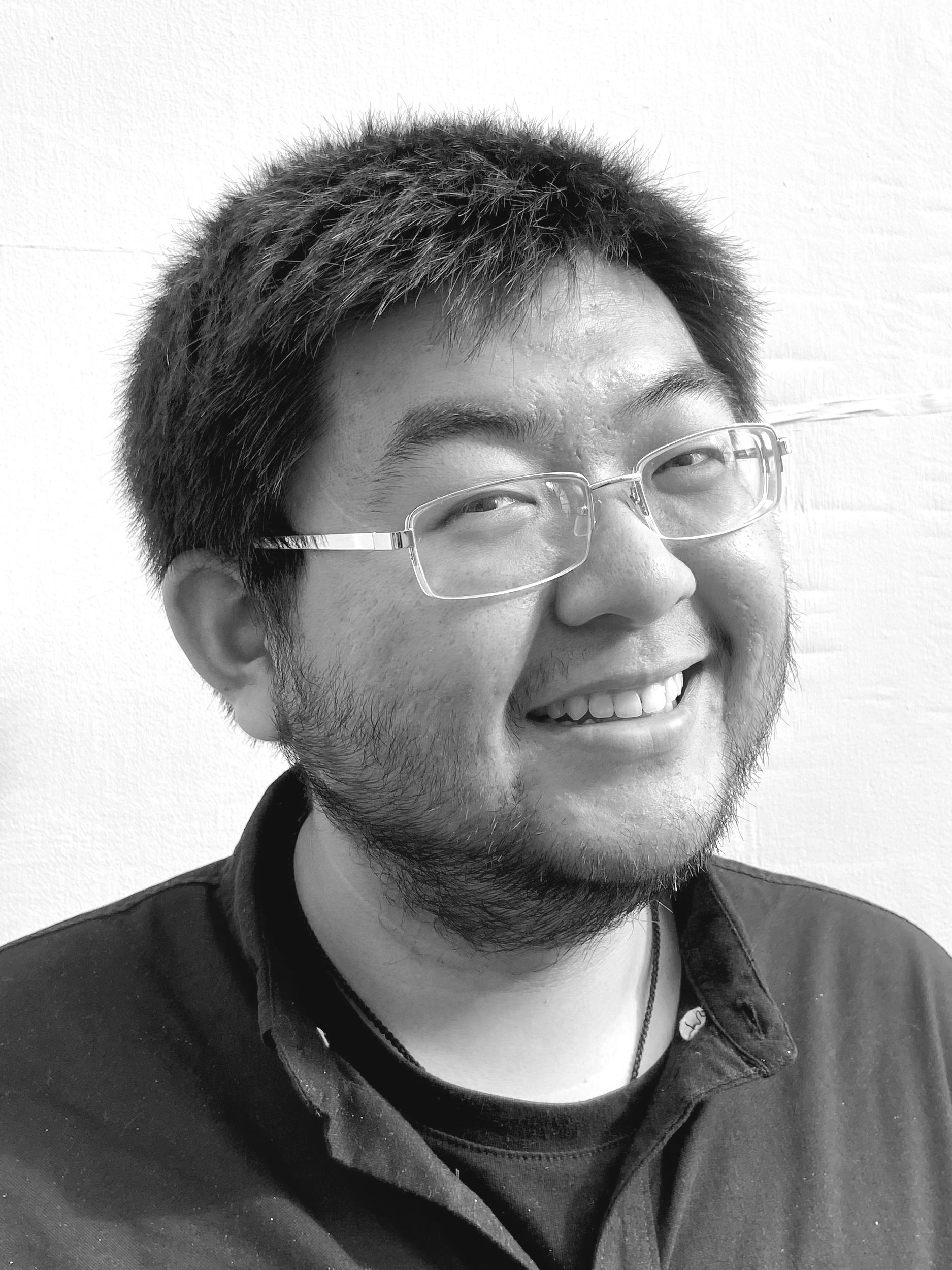}}]
    {Sen Yan (Graduate Student Member, IEEE)}
    received the bachelor’s degree in Internet of Things Engineering from Shandong University, in 2018 and the master’s degree in Computer Science - Intelligent Systems from Trinity College Dublin, the University of Dublin, in 2020. He is currently pursuing the Ph.D. degree in intelligent transportation systems at the School of Electronic Engineering, Dublin City University. Sen is working in the areas of data-driven methods used in intelligent transportation systems, especially shared micromobility systems. He is also interested in image processing and machine learning.
\end{IEEEbiography}

\vspace{-0.5\baselineskip}

\begin{IEEEbiography}[{\includegraphics[width=1in,height=1.33in,clip,keepaspectratio]{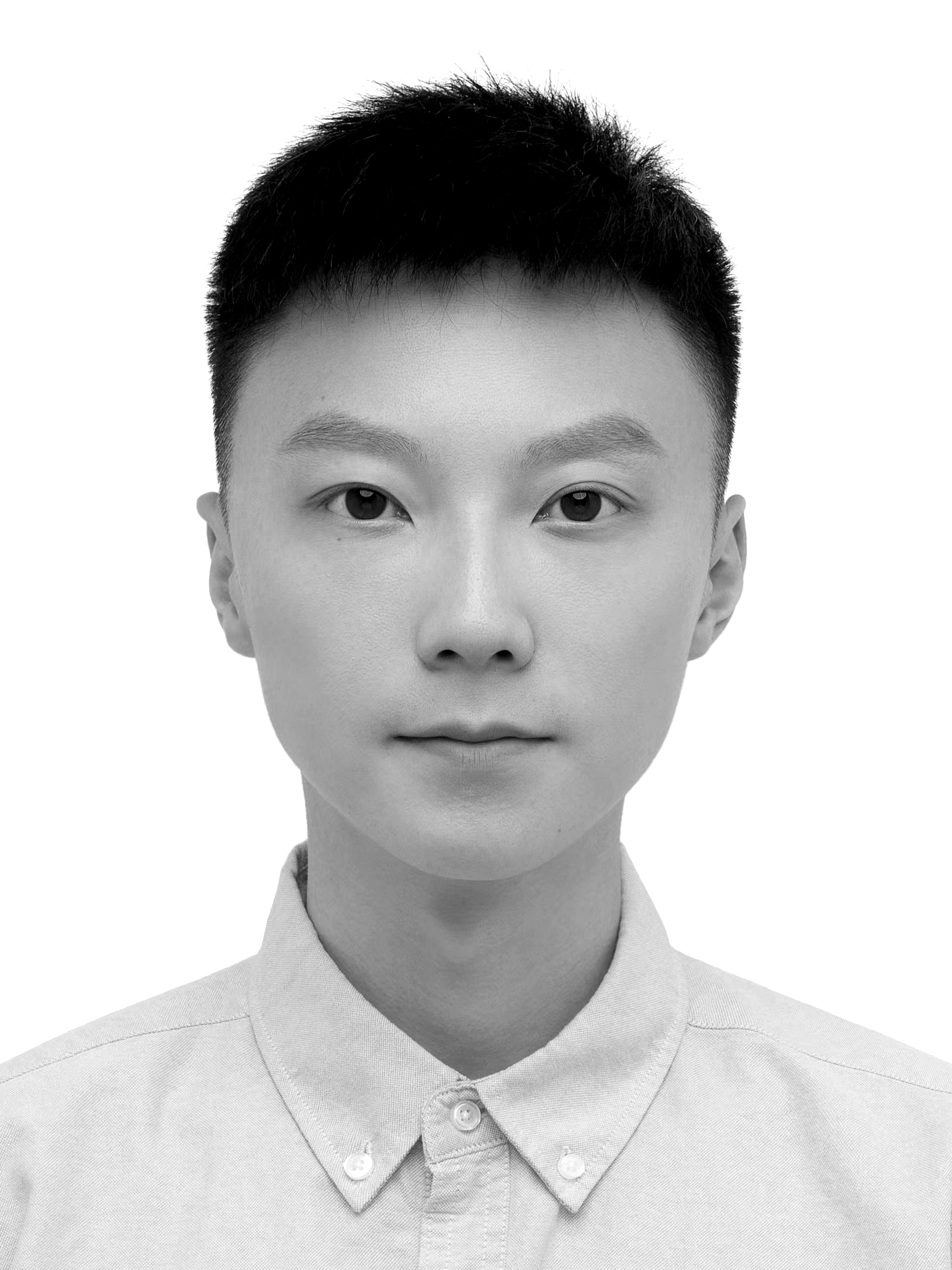}}]
    {Hongyuan Fang}
    received his Master's degree in Image Processing from Dublin City University (DCU), Dublin, Ireland, in 2022. He is currently pursuing his Ph.D. in the field of Intelligent Transportation at the School of Electronic Engineering, DCU. His research focuses on explainable anomaly detection in multivariate time series using deep learning techniques. His primary research interests include machine learning, graph neural networks, data-driven modeling, and computer vision. His work is particularly centered around the application of these technologies in intelligent transportation systems.
\end{IEEEbiography}

\vfill

\begin{IEEEbiography}[{\includegraphics[width=1in,height=1.33in,clip,keepaspectratio]{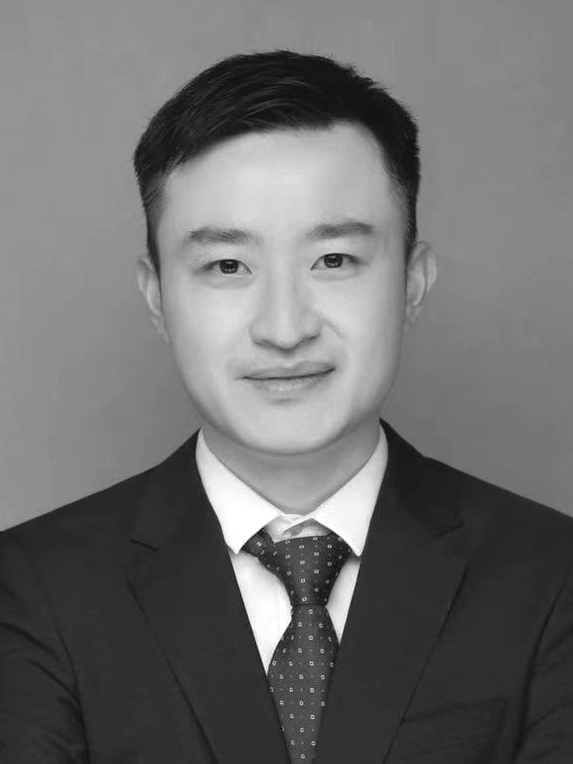}}]
    {Ji Li (Member, IEEE)}
     is awarded the Ph.D. degree in mechanical engineering from the University of Birmingham, UK, in 2020. He is a currently Research Fellow and works on the Connected and Autonomous Systems for Electrified Vehicles at Birmingham CASE Automotive Research \& Education Centre, University of Birmingham. He has authored 30 peer-reviewed international journal papers. His work has gained international recognition on many occasions including 1) Seal of Excellence from European Commission, and 2) three best paper awards at well-known international conferences. His focus covers three key domains in automotive engineering: multi-objective control, human-machine interaction, and cyber-physical systems, all oriented toward achieving cost-benefit and trustworthy engineering solutions.
\end{IEEEbiography}

\vspace{-2\baselineskip}

\begin{IEEEbiography}[{
    \includegraphics[width=1in,height=1.33in,clip,keepaspectratio]{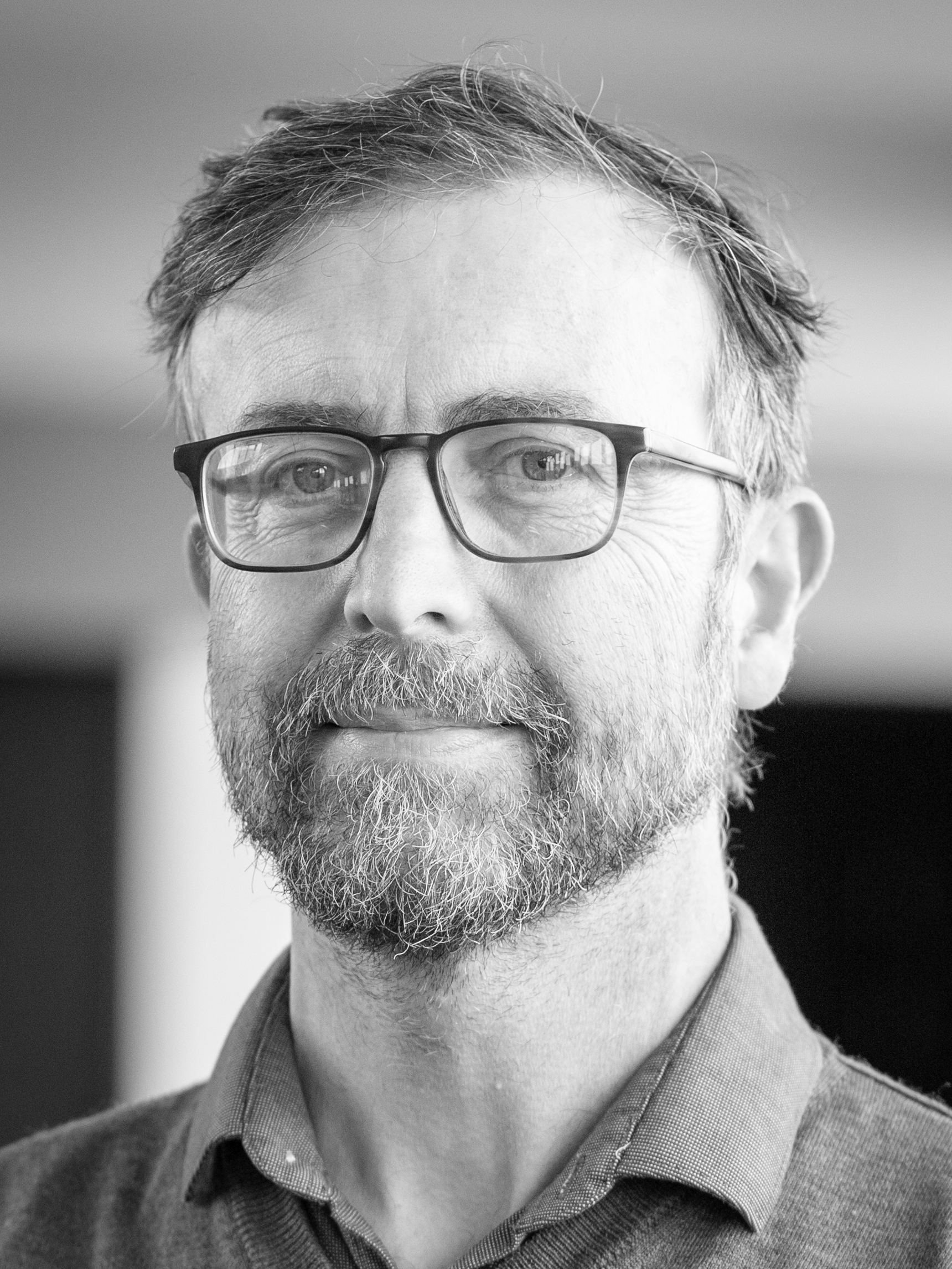}}]
    {Tomas Ward (Senior Member, IEEE)}
    received the B.E., M.Eng.Sc., and Ph.D. degrees in engineering from University College Dublin, Dublin, Ireland, in 1994, 1996, and 2000, respectively. He is currently a Full Professor and the AIB Chair of Data Analytics with the School of Computing, Dublin City University, Dublin. As a member of the Science Foundation Ireland-funded research center Insight, Ireland’s Data Analytics Research Centre, he studies how human health, performance, and decision-making can be better understood through new ways of sensing and interpreting our physiology and behavior.
\end{IEEEbiography}

\vspace{-2\baselineskip}

\begin{IEEEbiography}[{
    \includegraphics[width=1in,height=1.33in,clip,keepaspectratio]{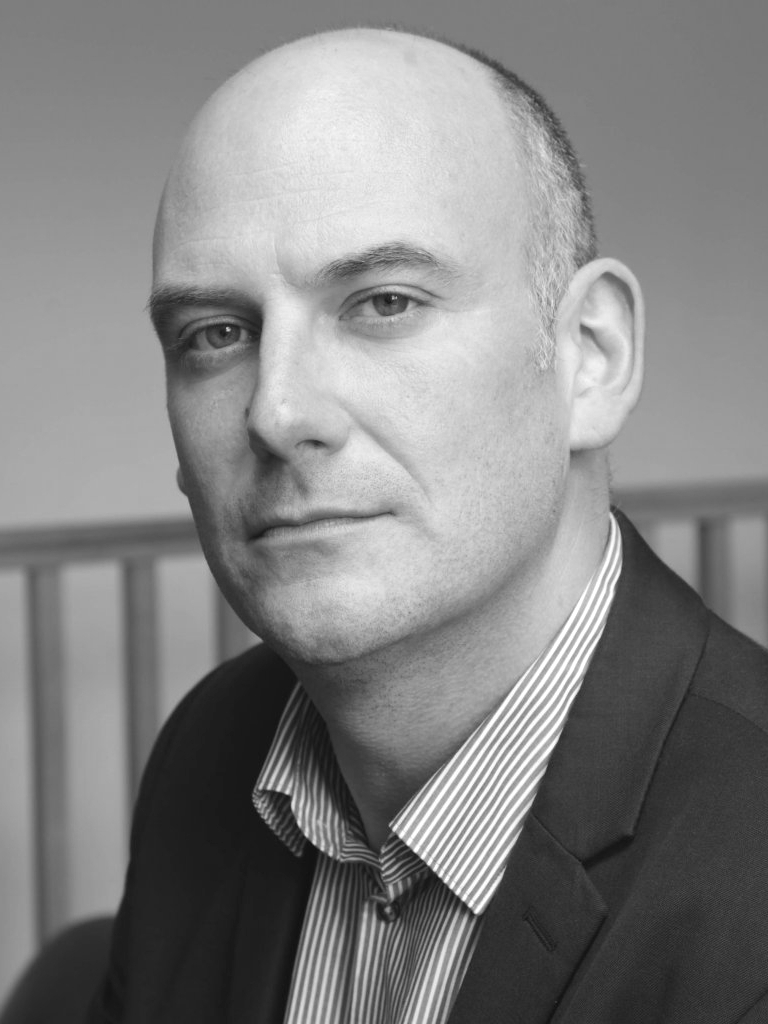}}]
    {Noel O'Connor (Member, IEEE)}
    was the Academic Director of DCU’s Research and Enterprise Hub on Information Technology and the Digital Society, with the responsibility of coordinating multi-disciplinary ICT-related research across the university. He is currently a Full Professor with the School of Electronic Engineering, Dublin City University (DCU), Ireland. He is also a CEO of the Insight SFI Research Centre for Data Analytics, Ireland’s largest SFI-funded research centre. Since 1999, he has published more than 400 peer-reviewed publications, made 11 standards submissions, and filed 7 patents. His research interests include multimedia content analysis, computer vision, machine learning, information fusion and multi-modal analysis for applications in security/safety, autonomous vehicles, medical imaging, the IoT and smart cities, multimedia content-based retrieval, and environmental monitoring. He is a member of the ACM. He is an Area Editor of Signal Processing: Image Communication (Elsevier) and an Associate Editor of the Journal of Image and Video Processing (Springer).
\end{IEEEbiography}

\vspace{-2\baselineskip}

\begin{IEEEbiography}[{
    \includegraphics[width=1in,height=1.33in,clip,keepaspectratio]{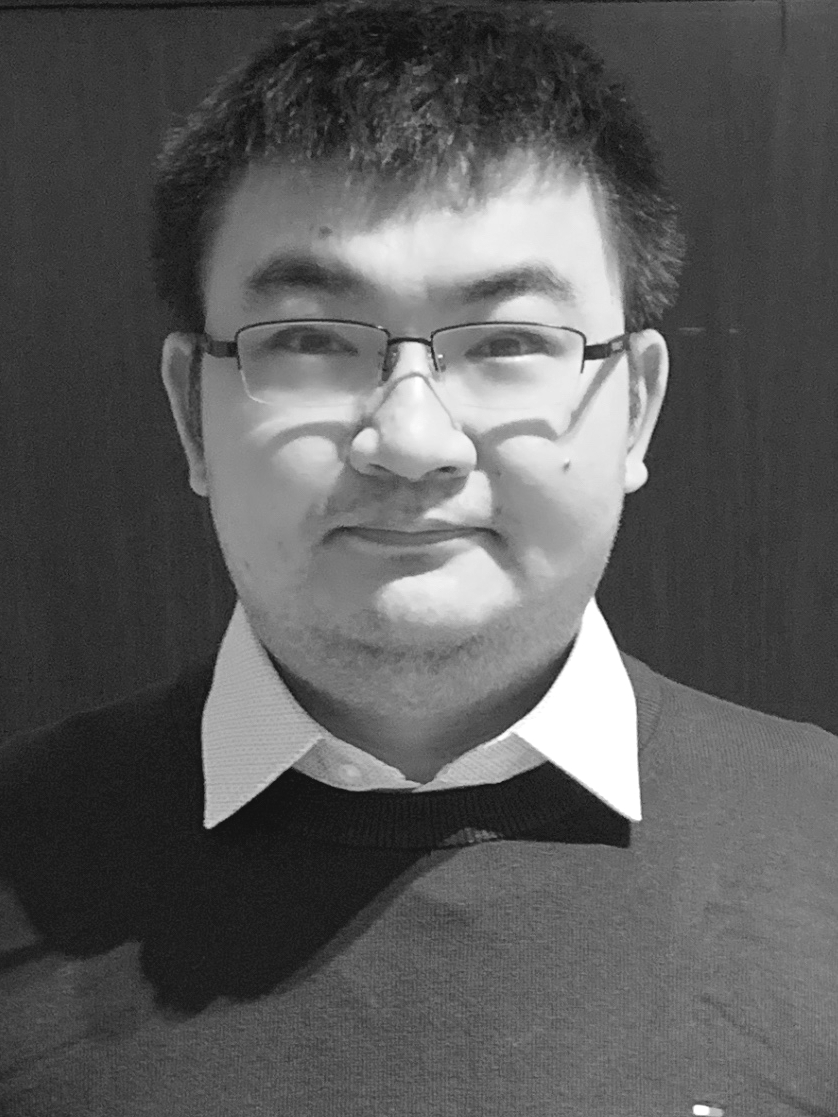}}]
    {Mingming Liu (Senior Member, IEEE)}
    is an Assistant Professor in the School of Electronic Engineering at Dublin City University (DCU). He is also affiliated with the SFI Insight Centre for Data Analytics as a Funded Investigator. He received the B. Eng (1st Hons) in Electronic Engineering from National University of Ireland Maynooth in 2011 and the PhD in Control Engineering from the Hamilton Institute at the same university in 2015. He is an IEEE senior member and has published over 50 papers to date, including ``IEEE Transactions on Smart Grids", ``IEEE Transactions on Intelligent Transportation Systems", ``IEEE Transactions on Automation Science and Engineering", "IEEE System Journal", ``IEEE Transactions on Transportation Electrification", ``Scientific Reports" and ``Automatica". Since 2018, he has secured more than 1.5 million euros in research funding as the independent PI. He is the management committee member in Ireland for EU COST Actions CA19126, CA20138 and CA21131. His research interests include control, optimization and machine learning with applications to 5G \& IoT, electric vehicles, smart grids, smart transportation, smart healthcare and smart cities.
\end{IEEEbiography}

\vfill